\documentclass[letterpaper]{article} % DO NOT CHANGE THIS
\usepackage{aaai2026}
\usepackage{times}  % DO NOT CHANGE THIS
\usepackage{helvet}  % DO NOT CHANGE THIS
\usepackage{courier}  % DO NOT CHANGE THIS
\usepackage[hyphens]{url}  % DO NOT CHANGE THIS
\usepackage{graphicx} % DO NOT CHANGE THIS
\usepackage{natbib}  % DO NOT CHANGE THIS AND DO NOT ADD ANY OPTIONS TO IT
\usepackage{caption} % DO NOT CHANGE THIS AND DO NOT ADD ANY OPTIONS TO IT
\usepackage{algorithm}
\usepackage{algorithmic}
\usepackage[table]{xcolor}
\usepackage{makecell}
\usepackage{array}
\usepackage{multirow}
\usepackage{bm}
\usepackage{amsfonts}
\usepackage{cancel}
\usepackage{xspace}
\newcommand{\ie}{\textit{i.e.}\xspace}
\usepackage{amsmath}    % 提供 split 环境
\usepackage{amssymb}   % 提供 \mathbb 符号
\definecolor{hrefpink}{rgb}{0.93,0.02,0.63}

\usepackage{newfloat}
\usepackage{listings}
\DeclareCaptionStyle{ruled}{labelfont=normalfont,labelsep=colon,strut=off} % DO NOT CHANGE THIS
\floatstyle{ruled}
\newfloat{listing}{tb}{lst}{}
\floatname{listing}{Listing}
\title{Robust Single-Stage Fully Sparse 3D Object Detection via Detachable Latent Diffusion}

\makeatletter
\renewcommand\thanks[1]{\footnotemark[\arabic{footnote}]\protected@xdef\@thanks{\@thanks
        \protect\footnotetext[\arabic{footnote}]{#1}}}
\makeatother

\author {
    Wentao Qu\textsuperscript{\rm 1},
    Guofeng Mei\textsuperscript{\rm 2},
    Jing Wang\textsuperscript{\rm 3},
    Yujiao Wu\textsuperscript{\rm 4},
    Xiaoshui Huang\textsuperscript{\rm 5}*,
    Liang Xiao\textsuperscript{\rm 1}*\thanks{\hspace{-6mm}*Corresponding Author. \textcolor{hrefpink}{https://github.com/QWTforGithub/RSDNet}}
}
\affiliations {
    \textsuperscript{\rm 1}NJUST, \textsuperscript{\rm 2}FBK, 
    \textsuperscript{\rm 3}HiDream.ai, 
    \textsuperscript{\rm 4}CSIRO, 
    \textsuperscript{\rm 5}SJTU \\
    quwentao@njust.edu.cn, gfmeiwhu@outlook.com, huangxiaoshui@163.com, xiaoliang@mail.njust.edu.cn
}
\usepackage{bibentry}
\begin{document}

\maketitle

\begin{abstract}
Denoising Diffusion Probabilistic Models (DDPMs) have shown success in robust 3D object detection tasks. 
Existing methods often rely on the score matching from 3D boxes or pre-trained diffusion priors. However, they typically require multi-step iterations in inference, which limits efficiency.
To address this, we propose a \textbf{R}obust single-stage fully \textbf{S}parse 3D object \textbf{D}etection \textbf{Net}work with a Detachable Latent Framework (DLF) of DDPMs, named RSDNet. Specifically, RSDNet learns the denoising process in latent feature spaces through lightweight denoising networks like multi-level denoising autoencoders (DAEs). This enables RSDNet to effectively understand scene distributions under multi-level perturbations, achieving robust and reliable detection. Meanwhile, we reformulate the noising and denoising mechanisms of DDPMs, enabling DLF to construct multi-type and multi-level noise samples and targets, enhancing RSDNet robustness to multiple perturbations. Furthermore, a semantic-geometric conditional guidance is introduced  to perceive the object boundaries and shapes, alleviating the center feature missing problem in sparse representations,  enabling RSDNet to perform in a fully sparse detection pipeline. Moreover, the detachable denoising network design of DLF enables RSDNet to perform single-step detection in inference, further enhancing detection efficiency.  Extensive experiments on public benchmarks show that RSDNet can outperform existing methods, achieving state-of-the-art detection. 
\end{abstract}

% Uncomment the following to link to your code, datasets, an extended version or similar.
%
% \begin{links}
%     \link{Code}{https://aaai.org/example/code}
%     \link{Datasets}{https://aaai.org/example/datasets}
%     \link{Extended version}{https://aaai.org/example/extended-version}
% \end{links}

\vspace{-15pt}
\section{Introduction}
Advances in 3D hardware and data synthesis technologies have made large-scale scene point clouds increasingly accessible. Reliably locating and recognizing targets in large-scale scenes, especially under various real-world noise, is crucial for real-time downstream tasks, such as autonomous driving \cite{mao20233d}, AR/VR \cite{li2025cross}, and robotics \cite{liu2024less}. Therefore, \textbf{\textit{robust and efficient 3D object detection}} has attracted increasing attention.

\begin{figure}[htp]
	\centering
\includegraphics[width=0.47\textwidth]
 {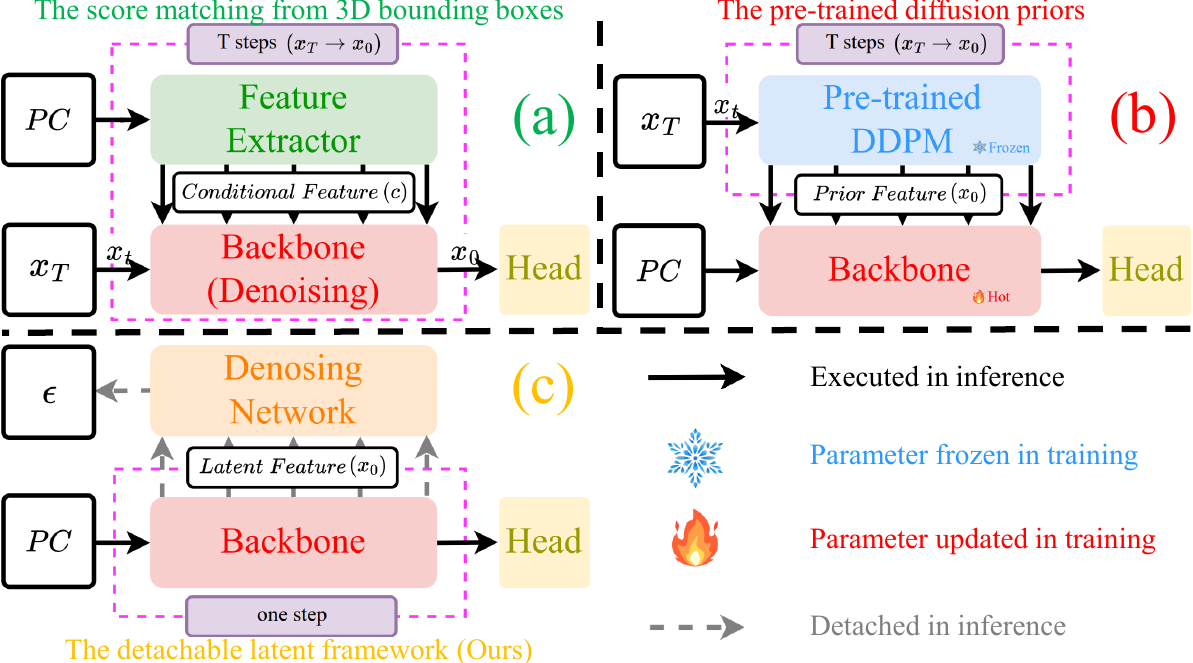}
 \vspace{-20pt}
\caption{Existing DDPM paradigms in 3D object detection: (a) performs denoising to estimate the box scores, then refines them via a detection head. (b) introduces a pre-trained diffusion prior to enhance detection accuracy. (c) (DLF) conducts the denoising learning via a denoising network. To generate high-quality 3D bounding boxes and feature priors, (a) and (b) demand multi-step inference. However, DLF can maintain the noise robustness while detaching the denoising network in inference, avoiding extra computational cost.}
\label{fig1}
\vspace{-6mm}
\end{figure}

In recent years, single-stage fully sparse 3D detection pipelines, built on hybrid architectures with 3D and 2D backbones, have become mainstream, exhibiting the impressive detection results \cite{fan2024fsd, zhang2024safdnet, zhang2025geobev, liu2025fshnet}. However, due to environmental or sensor factors, raw point clouds from 3D devices are often contaminated by multiple perturbations, such as point-level random noise and global geometric distortions (coordinate offset, scaling and rotation) \cite{rakotosaona2020pointcleannet, ding2024noise}. While these methods emphasize detection efficiency and accuracy, they often overlook robustness to perturbations, making them less reliable for stable performance in real-world scenarios~\cite{xiang2019generating}.

Along a research line, DDPMs \cite{ho2020denoising}, with a noise-robust architecture, have shown significant potential in robust object detection tasks \cite{ pellicer2024pudd, chu2025fire}. These methods typically estimate the scores from bounding boxes, refining them via a detection head (see Fig.~\ref{fig1}(a)) \cite{chen2023diffusiondet, ho2023diffusion, chen2024diffubox, ranasinghe2024monodiff}. Alternatively, they introduce the pre-trained diffusion priors into the  pipeline, improving the detection accuracy (see Fig.~\ref{fig1}(b)) \cite{xu20243difftection, zheng2024point}.

However, to obtain high-quality bounding boxes and feature priors, they inevitably require multi-step  iterations to accurately match the scores  in inference \cite{song2020denoising}. This requirement limits their practicality in downstream tasks with strict real-time constraints, such as autonomous driving.  Moreover, DDPMs that traditionally model Gaussian distributions struggle to be robust to other types of perturbations \cite{qu2024conditional,qu2025end}. 

These observations raise a central question:
\textbf{\textit{Can DDPMs overcome multi-step inference while modeling multiple perturbations?}} 
To answer this, we reveal the robustness source and rethink noising and denoising rules, providing new insights into DDPMs for 3D Object Detection (3DOD).

Inspired by these core insights, we design a Detachable Latent Framework (DLF, see Fig.~\ref{fig1}(c)) of DDPMs, effectively overcoming multi-step inference and preserving multiple noise robustness. Unlike Fig.~\ref{fig1}(a) and Fig.~\ref{fig1}(b), DLF guides the model to learn the denoising process via the denoising network in the latent feature space, but detached in inference. In this manner, DLF still inherits the DDPM training pattern, thus preserving noise robustness. Meanwhile, the detachable design for the denoising network avoids the multi-step inference. This also relaxes the score matching requirement, as the task result lies in the model backbone rather than the denoising network. Furthermore, we reformulate the noising and denoising mechanisms, enabling DLF to construct multi-type and multi-level noise samples and targets in training, thereby making the model robust to multiple perturbations. Moreover, DLF, supported by conditional guidance, can inject the task-specific knowledge priors for the model, enhancing the understanding for the task.

Furthermore, we propose a \textbf{R}obust single-step fully \textbf{S}parse 3D object \textbf{D}etection \textbf{Net}work based on DLF, called RSDNet. Specifically, RSDNet treats the denoising networks as multi-level denoising autoencoders (DAEs) \cite{xiang2023denoising, chen2024deconstructing}. This uses two lightweight denoising U-Nets ($<$6M), 3D Denoising U-Net (3DDU) and 2D Denoising U-Net (2DDU), guiding the 3D and 2D backbones to perform the denoising learning in a supervised manner. In this way, the backbones can understand the scene context in the multi-type and multi-level denoising learning, generating robust and generalized object-aware features for the detection head. Meanwhile, a semantic-geometric conditional guidance is injected to effectively mitigate the \textbf{\textit{center feature missing}} problem caused by downsampling or sparse convolution \cite{zhang2024safdnet}, enabling RSDNet to operate within an efficient fully sparse detection pipeline (3D and 2D sparse backbones). Moreover, thanks to the detachable design of DLF, RSDNet can achieve one-step inference, further improving the detection efficiency.

We lower the barrier, provide new insights, and encourage more researchers to further explore extensions of DDPMs in 3D tasks. Our key contributions can be summarized as:

\begin{itemize}
    \item We systematically reveal the robustness source and analyze the noising and denoising mechanisms, offering new knowledge for application of DDPMs in 3D tasks.
    \item We design a detachable latent framework of DDPMs, which can overcome multiple iterations in inference while preserving multiple noise robustness in training.
    \item We propose a robust single-step fully sparse detection network based on DLF, enabling one-step detection with strong robustness to multiple perturbations.
    \item Comprehensive experiments on public 3D detection benchmarks demonstrate that RSDNet can achieve strong robustness and significant detection performance.
   
\end{itemize}

\section{Related Works}

\noindent\textbf{Learnable 3D Object Detection.} Benefiting from  deep learning techniques, a lot of learnable 3D object detection methods have achieved success. Early methods typically employ dense voxels and multi-view 2D projections to establish ordered  3D representations within the detection pipeline \cite{chen2017multi, zhou2018voxelnet}. However, these methods either incur significant computational overhead due to numerous empty voxels or suffer from the loss of geometric details caused by object occlusion. PIXOR \cite{yang2018pixor} stands as the pioneer to convert point clouds into the 2D Bird’s-Eye View (BEV) representations, enabling 2D dense convolution effectively transferring to 3D object detection. Subsequently, some researchers propose coarse-to-fine detection pipelines based on BEV, further improving detection accuracy \cite{shi2019pointrcnn, yang2019std}. However, two-stage detection introduces significant cost in inference. For efficient detection pipelines, some methods implement single-stage detection via hybrid 3D sparse and 2D dense backbones \cite{bai2022transfusion, zhang2023hednet}. Recently, fully sparse pipelines show efficiency and effectiveness, further advancing 3D detection development \cite{zhang2024safdnet, liu2025fshnet}.

Although existing methods have demonstrated excellent detection performance, they overlook the fact that raw point clouds from 3D sensors are often perturbed. This makes them typically sensitive to noise, limiting their practical application. To address this issue, we introduce DDPMs into 3D object detection through a detachable latent framework. This guides the backbone to generate noise-robust features through the learning multi-level denoising process, while the detachability avoids extra computational cost in inference.

\noindent\textbf{DDPMs for Object Detection.} Some methods have explored DDPMs in robust detection tasks, exhibiting reliable and stable performance. DiffusionDet \cite{chen2023diffusiondet} marks the first integration of DDPMs as the fundamental mechanism for 2D object detection, displacing conventional query- and anchor-based paradigms. {This generates initial 2D bounding boxes via an iterative denoising process, followed by refinement of object locations and semantics using a detection head. Building upon this paradigm, subsequent methods have achieved notable improvements in the robustness and accuracy of 2D object detection \cite{chen2024camodiffusion, wang2024orienteddiffdet}.} Inspired by {the success in 2D detection}, 3D object detection methods have adapted DDPMs to learn the scores from 3D bounding boxes, achieving  robust and impressive performance \cite{ho2023diffusion, chen2024diffubox, ranasinghe2024monodiff}. {Furthermore, recent studies have explored leveraging DDPMs as pre-trained models for 3D object detection, yielding significant performance~\cite{xu20243difftection, zheng2024point}.}

However, to generate high-quality 3D bounding boxes and feature priors, these methods inevitably perform multi-step inference to accurately match the scores, limiting the real-world applicability. Moreover, global coordinate distortions also often exist in raw point clouds, {resulting in the unreliable detection}. Thus, we propose a robust single-stage fully sparse detection network base on DLF, conducting one-step inference, achieving multiple  perturbation robustness.

\section{Denoising Diffusion Probabilistic Models}

In this section, we first introduce the  background. Then, we explain the rationale behind multi-step inference and noise robustness. Subsequently, the noising and denoising mechanisms are reformulated to model multi-type perturbations.

\subsection{Background}
\label{sec:31}

Given an observed sample $\bm{c} \sim P_{sample}$, a fitting target $\bm{x_0} \sim P_{target}$, and a latent variable $\bm{x_T} \sim P_{noise}$, DDPMs achieve \textbf{\textit{the distribution transformation process between $P_{target}$ and $P_{noise}$}} via: a predefined diffusion process $q$ that gradually adds noise to $\bm{x_0}$ until $\bm{x_0}$ degrades into $\bm{x_T}$, and a trainable generation process $p_\theta$ that slowly removes perturbation from $\bm{x_T}$ until $\bm{x_T}$ recovers $\bm{x_0}$ conditioned on $\bm{c}$. 
Following this framework, DDPMs have  been successful in various 3D tasks \cite{qu2025end}. In 3DOD, Fig.~\ref{fig1}(a) ($\bm{c}\rightarrow$point cloud, $\bm{x_0}$$\rightarrow$3D box) and Fig.~\ref{fig1}(b) ($\bm{c}$$\rightarrow$task knowledge, $\bm{x_0}$$\rightarrow$point cloud) are commonly used paradigms. 

\begin{figure*}[htp]
	\centering
\includegraphics[width=\textwidth]
 {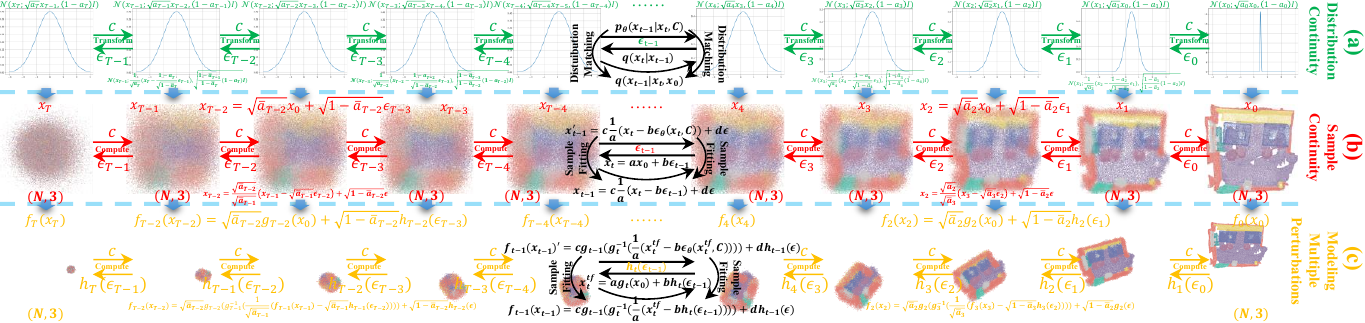}
\vspace{-18pt} 
\caption{Visualization of distribution matching, sample fitting, and modeling multiple perturbations. (a) The distribution matching requires each step to follow continuous and correlated distribution transformations, forming a complex formula chain \cite{ho2020denoising}. (b) The sample fitting focuses on effectively estimating the next-step sample, simplifying the construction conception \cite{bansal2023cold}. (c) The multi-type and multi-level noise samples and targets can be constructed using affine transformations. For example, $S_t(\bm{x_t}$$)=$$\sqrt{\overline{\alpha}_t}S_t(\bm{x_0}) +  \sqrt{1-\overline{\alpha}_t}S_t(\bm{\epsilon_{t-1}})$, this implements an $S_t$-fold scaling of $\bm{x_t}$.}\label{fig2}
    \vspace{-3mm} 
\end{figure*}

\noindent\textbf{Noising and Training Objective.} 

Given the strong performance observed in prior works \cite{ho2020denoising, qu2025end}, we adopt noise $\bm{\epsilon}$ as the fitting target, \ie, 
\vspace{-5pt}
\begin{equation}
\begin{split}
\label{f311}
 \bm{x_t}=\sqrt{\overline{\alpha}_t}\bm{x_0} + \sqrt{1-\overline{\alpha}_t}\bm{\epsilon_{t-1}}, \quad\quad\; \; \; \; \; \;\\
	L(\theta) =
 \mathbb{E}_{\bm{\epsilon_{t-1}} \sim \mathcal{N}(0,I)}||\bm{\epsilon_{t-1}} - \bm{\epsilon_\theta}(\bm{x_t},C,t)||_2^2,
\end{split}
\end{equation}
where $C=\{c_i|i=1,\cdots,S\}$ is an optional condition set ($C=\emptyset$ allowed) and $t\sim U[T]$ ($T{=}1000$ in this paper). The means that Eq.~\ref{f311} serves as a general objective for unconditional and conditional DDPMs~\cite{qu2024conditional}.

\noindent\textbf{The Gradient of The Data Distribution.} We can describe the objective using stochastic differential equations (SDEs):
\vspace{-10pt}
\begin{equation}
\begin{split}
	\label{f312}
	\alpha \bm{\epsilon_\theta}(\bm{x_t},C,t) = s_\theta\bm{(x_t},C,t) \approx \nabla_{\bm{x_t}} \log P_t(\bm{x_t}).
\end{split}
\end{equation}

Therefore, this noise objective is equivalent to the score  (\ie, the gradient of the data distribution)~\cite{song2021scorebased}, up to a constant factor $\alpha=-1/\sqrt{1-\overline{\alpha}_t}$.

\noindent\textbf{Denoising and Inference Sampling.} When approximating $\bm{\epsilon_\theta}(\bm{x_t},C,t)$ by $\bm{\epsilon_{t-1}}$, (\ie, $\bm{\epsilon_\theta}(\bm{x_t},C,t)\approx\bm{\epsilon_{t-1}}$), the trained $\bm{\epsilon_\theta}$ can then be used for iterative inference sampling as:
\vspace{-5pt}
\begin{equation}
\begin{split}
\label{f313}
	\bm{x_{t-1}}=\frac{1}{\sqrt{\alpha_t}}(\bm{x_t}-\frac{1-\alpha_t}{\sqrt{1-\overline{\alpha}_t}}\bm{\epsilon_{t-1}})\\
    +\sqrt{\frac{1-\overline{\alpha}_{t-1}}{1-\overline{\alpha}_t}(1-\alpha_t)} \bm{\epsilon}.\quad\;\;\;\;
\end{split}
\end{equation}

Eq.~\ref{f311} and Eq.~\ref{f313}  indicate  DDPMs adopt the sampling formulation $\bm{x_t}=\bm{\mu}_t+\sigma_t\bm{\epsilon}$ 
during both training and inference (\textbf{\textit{noising via $q(\bm{x_t}|\bm{x_0})$, denoising via $q(\bm{x_{t-1}}|\bm{x_t},\bm{x_0})$}}).

\subsection{DDPM Robustness Stems from the Training Stage}
\label{3.2}

We first explain why DDPMs require multi-step inference, and then analyze how their robustness to noise stems from the training phase rather than the inference procedure itself.

\noindent\textbf{Multi-Step Inference.} DDPMs actually construct richer samples and targets from  the modeled distribution in training than non-DDPMs. We provide a proof for this claim. For 3DOD, given a detection pair (point cloud $\bm{c}$, 3D box $\bm{x_0}$), a non-DDPM network $f_\theta$ can directly fit the input $\bm{c}$ to the target $\bm{x_0}$. However, the DDPM denoising network $\bm{\epsilon_\theta}$ takes $\bm{c}$ and $\{\bm{x_t}|t=1,\cdots,T\}$ as the inputs to fit the targets $\{\bm{\epsilon_{t-1}}|t=1,\cdots,T\}$  (assuming using the MSE loss):

\vspace{-8pt}
\begin{equation}
\begin{split}
\label{f321}
L_f({\theta}) =||\bm{x_0}-f_\theta(\bm{c})||_2^2,\quad \quad \quad\\
	L_\epsilon({\theta}) =\frac{1}{T}\sum_{t=1}^{T}||\bm{\epsilon_{t-1}}-\bm{\epsilon_\theta}(\bm{x_{t}}, \bm{c},t)||_2^2.
\end{split}
\end{equation}

As described in Eq.~\ref{f321}, under the same network architecture, DDPMs require $T$ times longer to fit the targets than non-DDPMs. \textbf{\textit{This is because transitioning between two distributions ($P_{target}$ and $P_{noise}$) with a large difference in one-step inference will lead to the significant error for the score matching}} \cite{song2021scorebased}.

\noindent\textbf{Robustness Source.} This noise construction pattern grants DDPMs adaptability to related distribution noise, as DDPMs can perceive more noise samples and targets in training than non-DDPMs. This also means that the DDPM robustness actually stems from \textbf{\textit{the noise samples and targets constructed in training}}.  Meanwhile, this training prior is independent of the inference approach. That is, DDPMs can \textbf{\textit{retain the noise robustness without following multi-step inference}}.

\subsection{DDPMs Can Model Multiple Perturbations}
\label{sec:33}

Besides point-level random noise, raw point clouds may also be affected by global geometric distortions, such as coordinate offset, scaling, rotation and other perturbations. In this paper, we rethink noising and denoising mechanisms, offering a way to model multiple perturbations in DDPMs.

\noindent\textbf{Distribution Matching.} In fact, $\bm{\epsilon_{t-1}}$ acts as a transition bridge between two distributions in Eq.~\ref{f311} (noising): $ p(\bm{x_t}|\bm{x_{t-1}})$ $\stackrel{\bm{\epsilon_{t-1}}}{\longleftarrow}$ $p(\bm{x_{t-1}}|\bm{x_{t-2}})$, and between two distributions in Eq.~\ref{f313} (denoising):
$ q(\bm{x_t}|\bm{x_{t+1}}, \bm{x_0})$  $\stackrel{\bm{\epsilon_{t-1}}}{\longrightarrow}$ $q(\bm{x_{t-1}}|\bm{x_t}, \bm{x_0})$. \textbf{\textit{When $\bm{\epsilon_\theta}$ accurately estimates $\bm{\epsilon_{t-1}}$, DDPMs actually achieve the distribution matching}}, \ie, $p_{\theta}(\bm{x_{t-1}}|\bm{x_t})$$\approx$$q(\bm{x_{t-1}}|\bm{x_t}, \bm{x_0})$ (see Fig.~\ref{fig2}(a)). This actually estimates the distribution of denoising at each step.

\noindent\textbf{Sample Fitting.} However, the distribution matching rule requires deriving $q(\bm{x_{t-1}}|\bm{x_t},\bm{x_0})$, involving a complex formula chain. This poses challenges when we remodel other distributions. Actually, $\bm{x_t}$ in training and inference are computed via $\bm{\mu_t}{+}\sigma_t\bm{\epsilon}$. Meanwhile, $\bm{\mu_t}$, $\sigma_t$, and $\bm{\epsilon}$ are all known in training. This means that the complex distribution matching task can actually be reinterpreted to a simple sample fitting problem: \textbf{\textit{When $\bm{\epsilon_\theta}$ accurately estimates $\bm{\epsilon_{t-1}}$, DDPMs actually achieve the sample fitting}}, \ie, $\bm{x'_{t-1}}$$\approx$$\bm{x_{t-1}}$ (see Fig.~\ref{fig2}(b)). In fact, $\bm{\epsilon_\theta}$ interacts only with noise samples and targets \textbf{\textit{without realizing the distribution concept}} in training. \textbf{\textit{The distribution transformation is manually imposed at inference.}}
Dropping the distribution concept, we can redefine the denoising using $q(\bm{x_{t-1}}|\bm{x_0})$ instead of $q(\bm{x_{t-1}}|\bm{x_t},\bm{x_0})$, decoupling the complex formula chain. 

\noindent\textbf{Modeling Multiple Perturbations.} Under the sample fitting rule, we can reformulate noising $q(\bm{x_t}|\bm{x_0})$ and denoising $q(\bm{x_{t-1}}|\bm{x_0})$  to apply \textbf{\textit{invertible affine transformations}}:

\begin{equation}\label{f331}
\begin{split}
 	\bm{x_t^{tf}}=ag_t(\bm{x_0})+b h_t(\bm{\epsilon_{t-1}}),\quad\quad\quad\quad\quad\\ \bm{x_{t{-}1}^{tf}}=cg_{t-1}(g_t^{-1}(\frac{1}{a}(\bm{x_t^{tf}}{-}bh_t(\bm{\epsilon_{t{-}1}})))){+}d h_{t{-}1}(\bm{\epsilon})
\end{split}
\end{equation}
where $a$, $b$, $c$, and $d$ are constant coefficients.  $f_t(\bm{x_t})$$=$$\bm{x_t^{tf}}$, $f_t(\cdot)$, $g_t(\cdot)$, $h_t(\cdot)$ denote affine transformation functions based on $t$ (derivations in Supplementary Material (SM)).

In fact, Eq.~\ref{f331} provides offers a general and flexible formulation to construct multi-type noise in DDPMs. $\bm{\epsilon}$ in Eq.~\ref{f331} can transcend the limitation of the Gaussian distribution \cite{austin2021structured}, even modeling the deterministic diffusion process \cite{bansal2023cold}, such as snowing and masking. 
% This means that we can construct 
This enables the construction of
any types of noise samples and targets without the distribution limitation, achieving robustness to multiple perturbations. Some works have actually adopted the sample fitting rule (\textbf{\textit{the noising follows $q(\bm{x_t}|\bm{x_0})$, the denoising follows $q(\bm{x_{t-1}}|\bm{x_0})$}}) \cite{bansal2023cold,naval2024ensembled}, but without the unified formulation. This way avoids complex denoising derivations. Moreover, the generation diversity still comes from $\bm{\epsilon}$.  \textbf{\textit{We also construct more types of noise samples and targets (see Fig.~\ref{fig2}(c), more implementations and visuals in SM)}}.

\section{Methodology}

\subsection{Detachable Latent Framework}

Based on the insights from Sec.~3.2 and Sec.~3.3, we propose a Detachable Latent Framework of DDPMs (DLF, see Fig.~\ref{fig1}(c)) to overcome multi-step inference and achieve multiple perturbation robustness. Unlike Fig.\ref{fig1}(a) and Fig.\ref{fig1}(b), DLF treats the denoising network $\bm{\epsilon_\theta}$ as an auxiliary branch, guiding the backbone $f_\psi$ to learn the multi-type and multi-level denoising process in latent feature spaces. 

Specifically, to learn the noise-robust denoising process, the latent feature $\bm{x_0^{lat}}$ from $f_\psi$ is perturbed to construct multi-type and multi-level noise samples and targets:

\vspace{-10pt}
\begin{equation}
\begin{split}
\label{f411}
f_t^*(\bm{x_t})=\sqrt{\overline{\alpha}_t}g_t^*(\bm{x_0^{lat}})+ \sqrt{1-\overline{\alpha}_t} \cdot h_t^*(\bm{\epsilon_{t-1}}),
\end{split}
\end{equation}
where $f_t^*(\cdot)$, $g_t^*(\cdot)$ and $h_t^*(\cdot)$ denote composite affine functions with intensity varying base on $t$, such as translation, scaling, rotation or other transformations (see Fig.~\ref{fig2}(c)).

Subsequently, the auxiliary denoising network $\bm{\epsilon_\theta}$ guides the backbone $f_\psi$ to learn the task-relevant information in multi-type and multi-level denoising learning:

\vspace{-10pt}
\begin{equation}
\begin{split}
\label{f412}
	L(\theta)=\mathbb{E}_{\bm{\epsilon_{t-1}} \sim \mathcal{N}(0,I)}||h_t^*(\bm{\epsilon_{t-1}})\\ - \bm{\epsilon_\theta}(f_t^*(\bm{x_t}),C_{task},t)||_2^2, \;\;
\end{split}
\end{equation}
where $C_{task}$ is task-specific conditions (knowledge priors).

Next, $f_\psi$ determines the task result in training and inference, while the denoise network $\bm{\epsilon_\theta}$ is detached in inference: 

\vspace{-10pt}
\begin{equation}
\begin{split}
\label{f413}
	L(\psi)=l_{task}(h_{task}(f_\psi(I_{task})), GT_{task}),\\ P_{task}=h_{task}(f_\psi(I_{task})),\quad \quad \quad
\end{split}
\end{equation}
where $I_{task}$, $GT_{task}$, $l_{task}(\cdot)$, $h_{task}(\cdot)$ and $P_{task}$ represent the task-related input, Ground Truth, loss function, task head and prediction, respectively.

This simple and effective DDPM paradigm:
\begin{itemize} 
    \item \textbf{Following Noise Construction Pattern.} Eq.~\ref{f411} indicates that DLF constructs multi-type and multi-level noise samples and targets in training, aligning with Eq.~\ref{f331} (noising), preserving the multi-type noise robustness.
    \item \textbf{Aligning with Training objective.} Eq.~\ref{f412} shows that DLF follows the original training objective (the score, $h_t^*(\bm{\epsilon_{t-1}})$$\approx$$\nabla_{\bm{x_t}} \log P_t(\bm{x_t})$, the noise target $h_t^*(\bm{\epsilon_{t-1}})$ performs better than $\bm{\epsilon_{t-1}}$, $\bm{x_0^{lat}}$ and $g_t^*(\bm{x_0^{lat}})$ in the additional ablation study of SM), aligning with Eq.~\ref{f312}, achieving the distribution matching continuity/the sample fitting continuity (see Sec.~3.3). 
    \item \textbf{Relaxing Score Matching Requirement.} Eq.~\ref{f413} presents that the task result relies on $f_\psi$ rather than $\bm{\epsilon_\theta}$, reducing the score learning difficulty in training.
    \item \textbf{With Minimal Cost.} Eq.~\ref{f411}, Eq.~\ref{f412} and Eq.~\ref{f413} mean that DLF introduces only limited cost in training by operating in latent feature spaces, while also avoiding extra training data and inference cost.
\end{itemize}

\begin{figure}[htp]
	\centering
\includegraphics[width=0.47\textwidth]
 {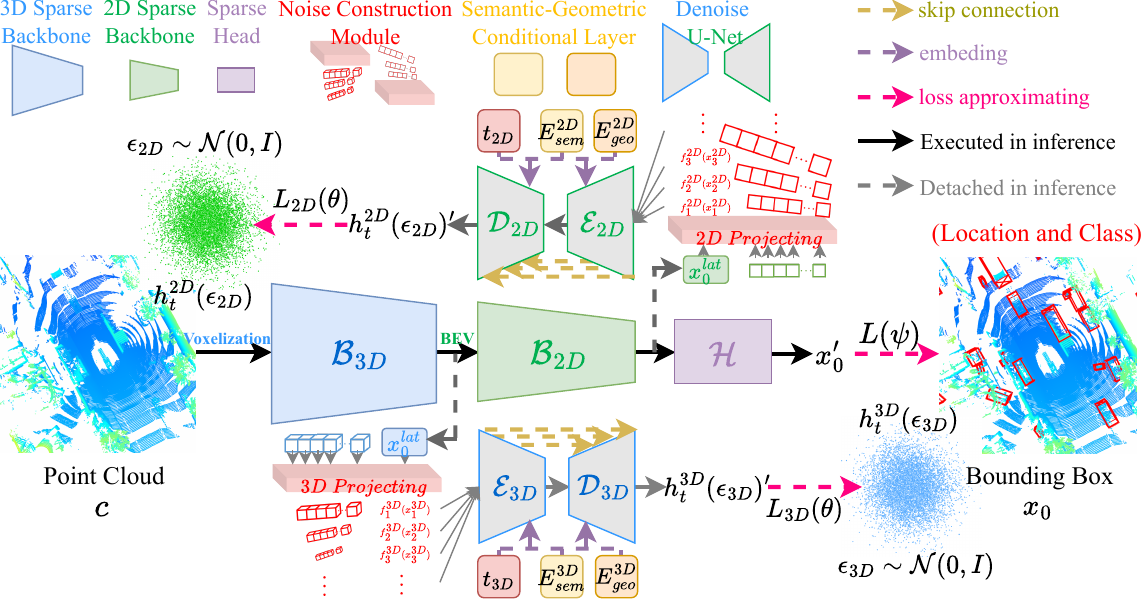}
 \vspace{-17pt}
\caption{The overall framework of RSDNet. DUNet takes multi-type and multi-level noise samples from NCM and semantic-geometric conditional embeddings from SGCL to guide FSP in denoising, while the DUNet detachable design  enables FSP to perform one-step detection in inference.}
	\label{fig3}
    \vspace{-5mm} 
\end{figure}

\subsection{Network Architecture}

In this section, we introduce the overall framework of RSDNet. This consists of four main components: the Fully Sparse Pipeline (FSP), the Noise Construction Module (NCM), the Semantic-Geometric Conditional Layer (SGCL) and the Denoising U-Net (DUNet), as illustrated in Fig.~\ref{fig3} (the implementation and parameter details in SM).

\noindent\textbf{Fully Sparse Pipeline.} FSP follows traditional hybrid sparse detection pipelines \cite{zhang2024safdnet, liu2025fshnet}, focusing on the detection results. The 3D sparse backbone first progressively (downsampling) extracts 3D sparse features via a voxel feature encoder (VFE) and five 3D sparse convolution layers. Then, two-stage linear self-attention block and a max-pooling layer are used to enhance the sparse representation perceive field.  Subsequently, the 2D sparse BEV representations from the compressed 3D sparse features further perceive contextual cues in the BEV space through a feature diffusion module and four 2D sparse convolution layers. Next, they are fed into the sparse detection head \cite{zhang2024safdnet} for the final prediction. We use only point clouds as inputs, ensuring FSP concentrates on  efficient and accurate object localization in a pure manner.

\noindent\textbf{Noise Construction Module.} NCM perturbs the latent feature $\bm{x_0^{lat}}$ from the backbone (2D or 3D), constructing multi-type and multi-level noise samples and targets. To ensure the lightweight design, $\bm{x_0^{lat}}$ is first progressively projected to a low-dimensional space through a three-layer sparse convolution. Then, as described in Eq.~\ref{f411}, NCM applies composite affine transformations to $\bm{x_0^{lat}}$ and $\bm{\epsilon_{t-1}}$:

\vspace{-10pt}
\begin{equation}
\begin{split}
\label{f421}
	g_t^*(\bm{x_0^{lat}})=R_t \cdot S_t(\bm{x_0^{lat}}-\bm{T_t}),\;\\
h_t^*(\bm{\epsilon_{t-1}})=R_t \cdot S_t(\bm{\epsilon_{t-1}}-\bm{T_t}),\\
\end{split}
\end{equation}
where $\bm{T_t}$, $S_t(\cdot)$ and $R_t$ denote the offset vector, the scaling function, and the rotation matrix  with intensity varying based on $t$, respectively (the implementation details in SM). 

Next, NCM performs the Gaussian noising process of DDPMs for $g_t^*({\bm{x_0^{lat}}})$ and $h_t^*({\bm{\epsilon_{t-1}}})$ to construct $f_t^*({\bm{x_t}})$.

\noindent\textbf{Semantic-Geometric Conditional Layer.} SGCL embeds semantic and geometric priors from the ground-truth 3D bounding boxes $B_{all}$, enhancing the backbone awareness for object boundaries and shapes in the denoising learning. This first selects the most relevant 3D bounding box $B_s$ for each point feature $F$ based on $\mathcal{L}_2$ distance between the box center $B_c$ and the voxel center $V_c$. Then, SGCL checks whether the voxel falls within the corresponding box:

\begin{figure}[htp]
	\centering
\includegraphics[width=0.47\textwidth]
 {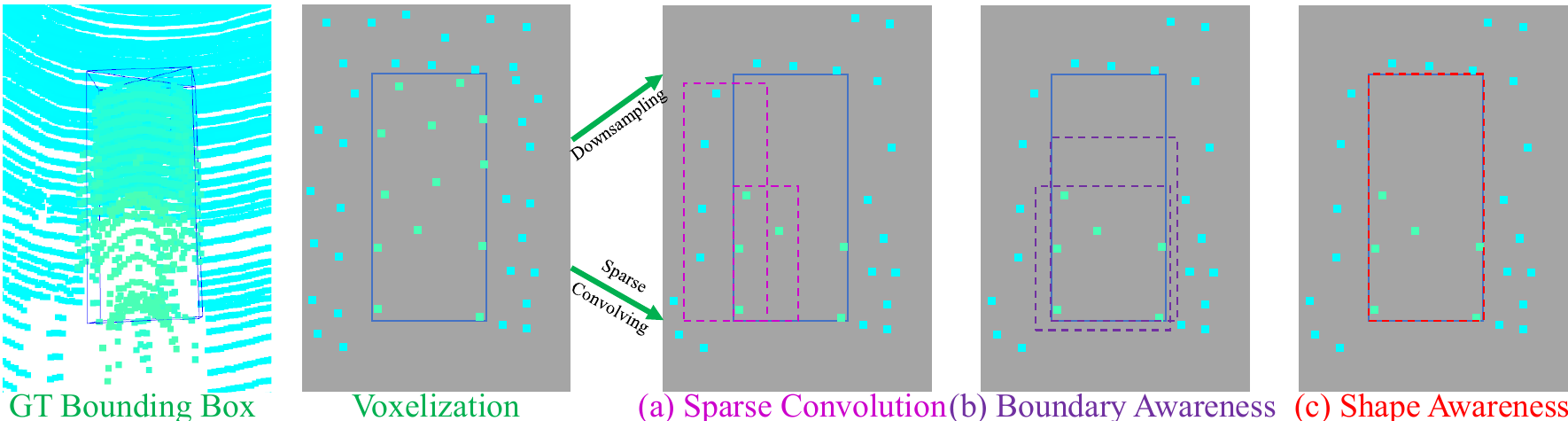}
 \vspace{-15pt}
\caption{The sparse features generated by downsampling or sparse convolution may lead to the center feature missing problem. (a) Missing center features may cause imprecise and unstable bounding box predictions. In such cases, background points may be misclassified as foreground due to the feature confusion, leading to false positives or incomplete detections. (b) The boundary aware guidance enhances the bounding box prediction for differentiating foreground points and background points. (c) The shape aware guidance further improves the alignment between the predicted bounding boxes and the ground truth bounding boxes.}
	\label{fig4}
    \vspace{-6mm} 
\end{figure}

\vspace{-10pt}
\begin{equation}
\begin{split}
\label{f422}
B_s=argmin(\mathcal{L}_2(B_c, V_c), B_{all}), \\
I_{mask}=\mathcal{I}(B_s, V_c), \quad\quad \;\;\;\;
\end{split}
\end{equation}
where $\mathcal{I}(\cdot)$ denotes an indicator function: the outputs 1 if the voxel falls within the corresponding box (foreground point), and 0 otherwise (background point). Meanwhile, SGCL uses 0 to fill in the box information for background voxels.

Subsequently, SGCL embeds the semantic embedding $E_{sem}$ from $I_{mask}$ and the geometric embedding $E_{geo}$ from $B_s$, and fuses them with the point feature $F$:

\vspace{-10pt}
\begin{equation}
\begin{split}
\label{f422}
	E_{sem}=embeding(I_{mask}), E_{geo}=mlp(B_f),\\
    F' = F + mlp(cat(E_{sem}, E_{geo})). \quad \quad \;\;\;
\end{split}
\end{equation}

The class label is excluded from $E_{geo}$ due to the absence of shape information about the object. Guided by the semantic and geometric priors, this can effectively mitigate the \textbf{\textit{center feature missing}} problem \cite{zhang2024safdnet} caused by downsampling or sparse convolution (see Fig.~\ref{fig4}).

\noindent\textbf{Denoise U-Net.} DUNet takes $f_t^*({\bm{x_t}})$ from NCM and performs the denoising learning guided by $E_{sem}$ and $E_{geo}$ from SGCL. For the hybrid backbone, DUNet includes two some architecture sub-networks: the 3D Denoising U-Net (3DDU) and 2D Denoising U-Net (2DDU). According to Sec.~4.1, DUNet should exhibit  lightweight due to the insignificant score matching requirement. Therefore, this follows a four-layer low-channel sparse convolution framework in the encoder and decoder \cite{shi2020points}. Meanwhile, DUNet  introduces the time label 
$t$ to model the diffusion process.

\subsection{Training and Inference}

\noindent\textbf{Training.} As mentioned in Sec.~4.1, the training objective of RSDNet includes the diffusion loss and the task loss:

% RSDNet should include the backbone loss and a simple denoising loss \cite{ho2020denoising}. Therefore, 

\vspace{-10pt}
\begin{equation}
\begin{split}
\label{f431}
	L_{total} = \lambda L(\theta) + L(\psi),
\end{split}
\end{equation}
where $L(\theta)=L_{3D}(\theta)+L_{2D}(\theta)$. Meanwhile, $L(\psi)=L_{reg}(\psi)+L_{cls}(\psi)$, $L_{reg}(\psi)$ and $L_{cls}(\psi)$ mean the regression
and classification loss functions (see Fig.~\ref{fig3}).

\noindent\textbf{Inference.} Thanks to the detachable design of DLF, RSDNet can perform detection in only one-step inference:

\vspace{-10pt}
\begin{equation}
\begin{split}
\label{f432}
	\bm{x'_0}=\mathcal{H}(\mathcal{B}_{2D}(\mathcal{B}_{3D}(\bm{c}))),
\end{split}
\end{equation}
where $\bm{x'_0}$ means the predicted bounding box (see Fig.~\ref{fig3}).

\section{Experiments}

\begin{table*}[h]
        %\scriptsize
	%\begin{center}
  \resizebox{1.0\textwidth}{!}{
	% \begin{tabular}{l|c|cccccccccccccccc}	
        \begin{tabular}{p{1.9cm}|p{1.0cm}p{1.0cm}|p{1.1cm}p{1.1cm}p{1.1cm}p{1.2cm}p{1.1cm}p{1.5cm}p{1.1cm}p{1.2cm}p{1.2cm}p{1.2cm}}	
        \Xhline{1pt}
  
        {Methods}
        &\makecell[c]{NDS}
        &\makecell[c]{mAP}
        &\makecell[c]{Car}
        &\makecell[c]{Truck}
        &\makecell[c]{Bus}
        &\makecell[c]{Trailer}
        &\makecell[c]{Vehicle}
        &\makecell[c]{Pedestrian}
        &\makecell[c]{Motor}
        &\makecell[c]{Bike}
        &\makecell[c]{Cone}
        &\makecell[c]{Barrier}\\
        
       \hline

        CenterPoint
        &\makecell[c]{66.5}
        &\makecell[c]{59.2}
        &\makecell[c]{84.9}        
        &\makecell[c]{57.4}
        &\makecell[c]{70.7}
        &\makecell[c]{38.1}
        &\makecell[c]{16.9}
        &\makecell[c]{85.1}
        &\makecell[c]{59.0}
        &\makecell[c]{42.0}
        &\makecell[c]{69.8}
        &\makecell[c]{68.3}\\

        PillarNeXt
        &\makecell[c]{68.4}
        &\makecell[c]{62.2}
        &\makecell[c]{85.0}        
        &\makecell[c]{57.4}
        &\makecell[c]{67.6}
        &\makecell[c]{35.6}
        &\makecell[c]{20.6}
        &\makecell[c]{86.8}
        &\makecell[c]{68.6}
        &\makecell[c]{53.1}
        &\makecell[c]{77.3}
        &\makecell[c]{69.7}\\

        VoxelNeXt\dag
        &\makecell[c]{68.7}
        &\makecell[c]{63.5}
        &\makecell[c]{83.9}        
        &\makecell[c]{55.5}
        &\makecell[c]{70.5}
        &\makecell[c]{38.1}
        &\makecell[c]{21.1}
        &\makecell[c]{84.6}
        &\makecell[c]{62.8}
        &\makecell[c]{50.0}
        &\makecell[c]{69.4}
        &\makecell[c]{69.4}\\

        HEDNet
        &\makecell[c]{71.4}
        &\makecell[c]{66.7}
        &\makecell[c]{87.7}        
        &\makecell[c]{60.6}
        &\makecell[c]{77.8}
        &\makecell[c]{50.7}
        &\makecell[c]{28.9}
        &\makecell[c]{87.1}
        &\makecell[c]{74.3}
        &\makecell[c]{56.8}
        &\makecell[c]{76.3}
        &\makecell[c]{66.9}\\

        FSDv2\dag
        &\makecell[c]{70.4}
        &\makecell[c]{64.7}
        &\makecell[c]{84.4}        
        &\makecell[c]{57.3}
        &\makecell[c]{75.9}
        &\makecell[c]{44.1}
        &\makecell[c]{28.5}
        &\makecell[c]{86.9}
        &\makecell[c]{69.5}
        &\makecell[c]{57.4}
        &\makecell[c]{72.9}
        &\makecell[c]{73.6}\\

        SAFDNet\dag
        &\makecell[c]{71.0}
        &\makecell[c]{66.3}
        &\makecell[c]{87.6}        
        &\makecell[c]{60.8}
        &\makecell[c]{78.0}
        &\makecell[c]{43.5}
        &\makecell[c]{26.6}
        &\makecell[c]{87.8}
        &\makecell[c]{75.5}
        &\makecell[c]{58.0}
        &\makecell[c]{75.0}
        &\makecell[c]{69.7}\\

        FSHNet\dag
        &\makecell[c]{71.7}
        &\makecell[c]{68.1}
        &\makecell[c]{88.7}        
        &\makecell[c]{61.4}
        &\makecell[c]{79.3}
        &\makecell[c]{47.8}
        &\makecell[c]{26.3}
        &\makecell[c]{89.3}
        &\makecell[c]{76.7}
        &\makecell[c]{60.5}
        &\makecell[c]{78.6}
        &\makecell[c]{72.3}\\

        Baseline\dag
        &\makecell[c]{71.2}
        &\makecell[c]{68.0}
        &\makecell[c]{87.7}        
        &\makecell[c]{62.1}
        &\makecell[c]{78.3}
        &\makecell[c]{42.7}
        &\makecell[c]{26.9}
        &\makecell[c]{88.7}
        &\makecell[c]{76.6}
        &\makecell[c]{59.5}
        &\makecell[c]{78.5}
        &\makecell[c]{79.2}\\

        \rowcolor{gray!20} 
        RSDNet\dag
        &\makecell[c]{71.9}
        &\makecell[c]{68.9}
        &\makecell[c]{88.4}        
        &\makecell[c]{63.1}
        &\makecell[c]{79.0}
        &\makecell[c]{43.2}
        &\makecell[c]{28.2}
        &\makecell[c]{89.2}
        &\makecell[c]{77.7}
        &\makecell[c]{59.9}
        &\makecell[c]{79.4}
        &\makecell[c]{80.4}\\

        \Xhline{1pt}
        
	\end{tabular}
	%\end{center}
 }
\vspace{-5pt} 
	\caption{The detection results on nuScenes. RSDNet significantly outperforms other methods in
 terms of NDS and mAP.}
	\label{tab521}
    \vspace{-1mm}
\end{table*}

\begin{table*}[h]
        %\scriptsize
	%\begin{center}
  \resizebox{1.0\textwidth}{!}{
	% \begin{tabular}{l|c|cccccccccccccccc}	
        \begin{tabular}{p{1.9cm}|p{1.8cm}p{0.005cm} p{1.8cm}p{0.005cm}p{1.4cm}p{1.5cm}p{1.4cm}p{0.005cm}p{1.4cm}p{1.5cm}p{1.4cm}}	
        \Xhline{1pt}
  
        \multirow{2}{*}{Methods}
        &\makecell[c]{LEVEL1}
        &\quad
        &\makecell[c]{LEVEL2}
        &\quad
        &\multicolumn{3}{c}{LEVEL1}
        &\quad
        &\multicolumn{3}{c}{LEVEL2}\\
        \cline{2-2}
        \cline{4-4}
        \cline{6-8} 
        \cline{10-12}
        
        &\makecell[c]{mAP/mAPH}
        &\quad
        &\makecell[c]{mAP/mAPH}
        &\quad
        &\makecell[c]{Vehicle}
        &\makecell[c]{Pedestrian}        
        &\makecell[c]{Cyclist}
        &\quad
        &\makecell[c]{Vehicle}
        &\makecell[c]{Pedestrian}        
        &\makecell[c]{Cyclist}\\
        
       \hline

        CentorPoint
        &\makecell[c]{74.4/71.7}
        &\quad
        &\makecell[c]{68.2/65.8}
        &\quad
        &\makecell[c]{74.2/73.6}
        &\makecell[c]{76.6/70.5}      
        &\makecell[c]{72.3/71.1}
        &\quad
        &\makecell[c]{66.2/65.7}
        &\makecell[c]{68.8/63.2}      
        &\makecell[c]{69.7/68.5}\\

        PillarNeXt
        &\makecell[c]{78.0/75.7}
        &\quad
        &\makecell[c]{71.9/69.7}
        &\quad
        &\makecell[c]{78.4/77.9}
        &\makecell[c]{82.5/77.1}      
        &\makecell[c]{73.2/72.2}
        &\quad
        &\makecell[c]{70.3/69.8}
        &\makecell[c]{74.9/69.8}      
        &\makecell[c]{70.6/69.6}\\

        VoxelNeXt\dag
        &\makecell[c]{78.6/76.3}
        &\quad
        &\makecell[c]{72.2/70.1}
        &\quad
        &\makecell[c]{78.2/77.7}
        &\makecell[c]{81.5/76.3}      
        &\makecell[c]{76.1/74.9}
        &\quad
        &\makecell[c]{69.9/69.4}
        &\makecell[c]{73.5/68.6}      
        &\makecell[c]{73.3/72.2}\\

        HEDNet
        &\makecell[c]{81.4/79.4}
        &\quad
        &\makecell[c]{75.3/73.4}
        &\quad
        &\makecell[c]{81.1/80.6}
        &\makecell[c]{84.4/80.0}      
        &\makecell[c]{78.7/77.7}
        &\quad
        &\makecell[c]{73.2/72.7}
        &\makecell[c]{76.8/72.6}      
        &\makecell[c]{75.8/74.9}\\

        FSDv2\dag
        &\makecell[c]{80.3/79.5}
        &\quad
        &\makecell[c]{75.6/73.5}
        &\quad
        &\makecell[c]{79.8/79.3}
        &\makecell[c]{84.8/79.7}      
        &\makecell[c]{80.7/79.6}
        &\quad
        &\makecell[c]{71.4/71.0}
        &\makecell[c]{77.4/72.5}      
        &\makecell[c]{77.9/76.8}\\

        SAFDNet\dag
        &\makecell[c]{81.8/79.8}
        &\quad
        &\makecell[c]{75.7/73.9}
        &\quad
        &\makecell[c]{80.6/80.1}
        &\makecell[c]{84.7/80.4}      
        &\makecell[c]{80.0/79.0}
        &\quad
        &\makecell[c]{72.7/72.3}
        &\makecell[c]{77.3/73.1}      
        &\makecell[c]{77.2/76.2}\\

        FSHNet\dag
        &\makecell[c]{82.7/80.6}
        &\quad
        &\makecell[c]{77.1/74.9}
        &\quad
        &\makecell[c]{82.2/81.7}
        &\makecell[c]{85.9/80.8}      
        &\makecell[c]{80.5/79.4}
        &\quad
        &\makecell[c]{74.5/74.0}
        &\makecell[c]{78.9/73.9}      
        &\makecell[c]{78.0/76.9}\\

        Baseline\dag
        &\makecell[c]{82.7/80.5}
        &\quad
        &\makecell[c]{76.9/74.8}
        &\quad
        &\makecell[c]{82.0/81.5}
        &\makecell[c]{85.7/80.7}      
        &\makecell[c]{80.3/79.2}
        &\quad
        &\makecell[c]{74.2/73.8}
        &\makecell[c]{78.8/73.7}       
        &\makecell[c]{77.8/76.8}\\

        \rowcolor{gray!20} 
        RSDNet\dag
        &\makecell[c]{83.7/81.4}
        &\quad
        &\makecell[c]{77.8/75.6}
        &\quad
        &\makecell[c]{82.8/82.3}
        &\makecell[c]{86.7/81.5}      
        &\makecell[c]{81.6/80.5}
        &\quad
        &\makecell[c]{74.9/74.5}
        &\makecell[c]{79.8/74.7}       
        &\makecell[c]{78.7/77.7}\\

        \Xhline{1pt}
        
	\end{tabular}
	%\end{center}
 }
 \vspace{-5pt}
	\caption{The detection results on Waymo Open. RSDNet demonstrates excellent detection performance in large-scale scenes. }
	\label{tab522}
    \vspace{-5mm}
\end{table*}

\subsection{Experiment Setup}

\textbf{Dataset.} We perform the main experiments on nuScenes \cite{caesar2020nuscenes}, following the official protocol to divide train/val/test with 700/150/150 scenes. We also conduct experiments on Waymo Open \cite{sun2020scalability},  splitting
train/val/test into 798/202/150 scenes.

\noindent\textbf{Detection Methods.} We compare RSDNet with current popular detection methods: CenterPoint \cite{yin2021center}, PillarNeXt \cite{li2023pillarnext}, VoxelNeXt\dag \cite{chen2023voxelnext}, HEDNet \cite{zhang2023hednet}, FSDv2\dag \cite{fan2024fsd}, SAFDNet\dag \cite{zhang2024safdnet}, FSHNet\dag \cite{liu2025fshnet}. \dag means a fully sparse detector.

\noindent\textbf{Baseline.} To validate the effectiveness of our method, we remove 3DDU and 2DDU from RSDNet and treat the Fully Sparse  Pipeline (FSP) as the baseline.

\subsection{Comparison of Detection Results}

\begin{figure*}[htp]
	\centering
\includegraphics[width=\textwidth]
 {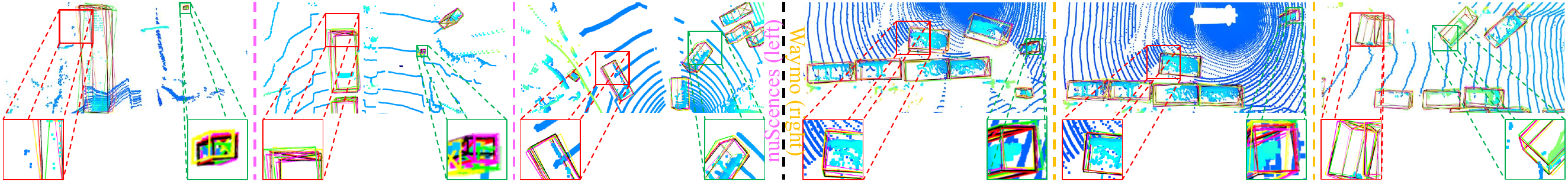}
\vspace{-15pt} 
\caption{The visualizations on nuScenes (left) and Waymo Open (right). The red, yellow, green, pink, and black boxes represent the Ground Truth, HEDNet, SAFDNet, FSHNet, and RSDNet, respectively. RSDNet shows better performance in small objects. }\label{fig5}
    \vspace{-3mm} 
\end{figure*}

\textbf{Results on nuScenes.} We first conduct the evaluation on the nuScenes dataset. As shown in Tab.~\ref{tab521}, RSDNet achieves the significant detection performance, outperforming the existing state-of-the-art methods. This is because, multi-type and multi-level denoising learning improves the robustness and generalization of RSDNet in unseen scenes, especially for noise-sensitive small objects. Meanwhile, the  semantic-geometric conditional guidance alleviates the center feature missing problem from downsampling and sparse convolution, further enhancing small object detection results. Fig.~\ref{fig5}(left) further demonstrates the reliable and significant detection results of RSDNet for small objects.

\noindent\textbf{Results on Waymo Open.} Furthermore, we also conduct the evaluation on the longer-range Waymo Open. As shown in Tab.~\ref{tab522}, RSDNet still achieves state-of-the-art detection performance. Benefiting from the robustness to perturbations, RSDNet exhibits strong generalization in  scenes with the sparser object distribution \cite{qu2025end}. Fig.~\ref{fig5}(right) shows the superior performance in large-scale scenes.

\subsection{Validation for Perturbation Robustness}

To verify reliable and stable detection, we conduct the robustness evaluation for multiple perturbations on nuScenes.

\noindent\textbf{Point-level Random Noise.} Raw point clouds from the sensor often contain point-wise random noise. We add the Gaussian noise $\bm{n_G} \sim \mathcal{N}(\bm{n_G};\bm{0},\tau\bm{I})$ to the normalized input of the model \cite{qu2025end}, \ie, $\bm{c'}=\bm{c} + \bm{n_G}$, evaluating the  robustness for point-level random noise. Tab.~\ref{tab531} shows that RSDNet exhibits the strong noise robustness compared to other methods. Meanwhile, the detection results drop close to 0 when $\tau$=0.15, indicating that existing detection methods are relatively sensitive to point-level random noise.

\vspace{-5pt}
\begin{table}[h]
        \scriptsize
	%\begin{center}
  \resizebox{0.475\textwidth}{!}{
	\begin{tabular}{p{1.2cm}p{0.8cm}p{0.8cm}p{0.8cm}p{0.005cm}p{0.8cm}p{0.9cm}p{0.8cm}}	
        \Xhline{1pt}
  
        \multirow{2}{*}{Methods}
        &\multicolumn{3}{c}{Small $\tau$ (mAP)} 
        &\quad
        &\multicolumn{3}{c}{Big $\tau$ (mAP) } \\
         \cline{2-4} \cline{6-8}
        
        &\makecell[c]{$\bm{\tau}$=0.01}
        &\makecell[c]{$\bm{\tau}$=0.05}
        &\makecell[c]{$\bm{\tau}$=0.08}
        &\quad
        &\makecell[c]{$\bm{\tau}$=0.10}
        &\makecell[c]{$\bm{\tau}$=0.125}
        &\makecell[c]{$\bm{\tau}$=0.15}\\
        
        \Xhline{1pt}

        HEDNet
        &\makecell[c]{66.6}
        &\makecell[c]{58.5}
        &\makecell[c]{39.7}
        &\quad
        &\makecell[c]{20.4}
        &\makecell[c]{5.7}
        &\makecell[c]{1.0}\\

        SAFDNet\dag
        &\makecell[c]{66.5}
        &\makecell[c]{57.4}
        &\makecell[c]{38.7}
        &\quad
        &\makecell[c]{20.2}
        &\makecell[c]{4.4}
        &\makecell[c]{1.0}\\

        FSHNet\dag
        &\makecell[c]{67.9}
        &\makecell[c]{60.1}
        &\makecell[c]{40.4}
        &\quad
        &\makecell[c]{23.5}
        &\makecell[c]{9.4}
        &\makecell[c]{1.2}\\

        Baseline\dag
        &\makecell[c]{67.8}
        &\makecell[c]{59.8}
        &\makecell[c]{40.2}
        &\quad
        &\makecell[c]{23.0}
        &\makecell[c]{8.9}
        &\makecell[c]{1.2}\\

        \rowcolor{gray!20} 
        RSDNet\dag
        &\makecell[c]{68.7}
        &\makecell[c]{64.0}
        &\makecell[c]{51.2}
        &\quad
        &\makecell[c]{34.2}
        &\makecell[c]{18.4}
        &\makecell[c]{7.5}\\
        \Xhline{1pt}
        
	\end{tabular}
	%\end{center}
 }
 \vspace{-5pt}
	\caption{The results for Gaussian noise on nuScenes. RSDNet shows excellent robustness to point-level random noise.}
	\label{tab531}
    \vspace{-3mm}
\end{table}

\begin{figure}[htp]
	\centering
\includegraphics[width=0.47\textwidth]
 {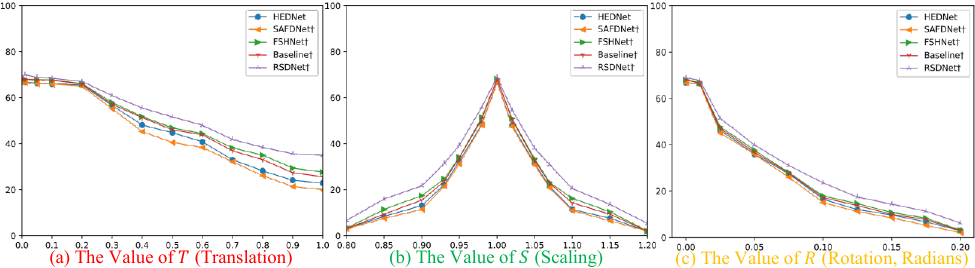}
 \vspace{-15pt}
\caption{(a), (b), and (c) mean the results of applying translation, scale, and rotation perturbations, respectively.}
	\label{fig6}
    \vspace{-5mm} 
\end{figure}

\noindent\textbf{Global Geometric Distortions.} Coordinate offsets, scaling, and rotations may also often be presented in raw point clouds. Similarly, we apply translation ($\bm{c'}=\bm{c} - \bm{T}$), scaling ($\bm{c'}=S\bm{c}$), and rotation ($\bm{c'}=R \cdot \bm{c}$) perturbations to the normalized input of the model. Fig.~\ref{fig6} shows the results. Benefiting from multi-type noise samples and targets, RSDNet exhibits strong robustness to global geometric distortions.

\subsection{Generalization for DLF}

As mentioned  in Sec.~4.1, DLF serves as a framework of DDPMs for 3D tasks. Therefore, we further conduct the generalization experiments for DLF (the details in SM).

\noindent\textbf{Other Backbones.}
We first conduct the experiments for
introducing DLF into HEDNet and SAFDNet. We simply integrate 3DDU and 2DDU into HEDNet and SAFDNet. Tab.~\ref{tab541} demonstrates that integrating DLF obtains the better  noise robustness and detection performance for HEDNet and SAFDNet. Meanwhile, this avoids introducing the additional inference cost, as the detachable design of DLF.

\vspace{-5pt}
\begin{table}[h]
        \scriptsize
        \resizebox{0.475\textwidth}{!}{
%	\begin{center}
	\begin{tabular}{p{1.8cm}p{1.0cm}p{0.7cm}p{0.7cm}p{0.8cm}p{0.8cm}p{0.8cm}p{1.1cm}}	

        \Xhline{1pt}
        \makecell[l]{Methods}
        &\makecell[c]{$\#$Params}
        &\makecell[c]{NDS}
        &\makecell[c]{mAP}
        &\makecell[c]{$\tau$=0.01}
        &\makecell[c]{$\tau$=0.05}
        &\makecell[c]{$\tau$=0.10}
        &\makecell[c]{MIT/MM}\\
        \hline

        \makecell[l]{HEDNet}
        &\makecell[c]{15.3M}
        &\makecell[c]{71.1}
        &\makecell[c]{66.8}
        &\makecell[c]{66.6}
        &\makecell[c]{58.5}
        &\makecell[c]{20.4}
        &\makecell[c]{0.18s/3.8G}\\

        \rowcolor{gray!20} 
        \makecell[l]{HEDNet+DLF}
        &\makecell[c]{18.3M}
        &\makecell[c]{71.3}
        &\makecell[c]{67.3}
        &\makecell[c]{67.1}
        &\makecell[c]{60.8}
        &\makecell[c]{31.1}
        &\makecell[c]{0.18s/3.8G}\\

        \makecell[l]{SAFDNet\dag}
        &\makecell[c]{15.7M}
        &\makecell[c]{69.9}
        &\makecell[c]{66.7}
        &\makecell[c]{66.5}
        &\makecell[c]{57.4}
        &\makecell[c]{20.2}
        &\makecell[c]{0.15s/4.7G}\\

        \rowcolor{gray!20} 
        \makecell[l]{SAFDNet+DLF\dag}
        &\makecell[c]{18.7M}
        &\makecell[c]{71.4}
        &\makecell[c]{67.4}
        &\makecell[c]{67.3}
        &\makecell[c]{60.5}
        &\makecell[c]{30.8}
        &\makecell[c]{0.15s/4.7G}\\

        \rowcolor{gray!20} 
        \makecell[l]{RSDNet\dag}
        &\makecell[c]{16.5M}
        &\makecell[c]{71.9}
        &\makecell[c]{68.9}
        &\makecell[c]{68.7}
        &\makecell[c]{64.0}
        &\makecell[c]{34.2}
        &\makecell[c]{0.16s/5.8G}\\

        \Xhline{1pt}

	\end{tabular}
	%\end{center}
 }
  \vspace{-5pt} 
	\caption{The results of multiple backbones on nuScenes. `MIT' and `MM' indicate the \textit{Mean Inference Time} and the \textit{Mean Memory} for each point cloud, respectively. We run on an  NVIDIA 3090 GPU with batch size=1, workers=1.}
	\label{tab541}
    \vspace{-3mm}
\end{table}

\noindent\textbf{Other 3D Tasks.} We further extend DLF to classification tasks. This uses PointNet and PointNet++ as backbones, and implements DUNet based on an additional PointNet++. As shown in Tab.~\ref{tab542}, incorporating DLF  significantly improves the robustness and classification accuracy of backbones.

\vspace{-5pt}
\begin{table}[h]
        \scriptsize
	%\begin{center}
  \resizebox{0.475\textwidth}{!}{
	\begin{tabular}{p{1.4cm}p{0.5cm}p{0.5cm}p{0.8cm}p{0.7cm}p{0.005cm}p{0.5cm}p{0.5cm}p{0.8cm}p{0.7cm}}	
        \Xhline{1pt}
        
        \multirow{2}{*}{Methods}
        &\multicolumn{4}{c}{PointNet \cite{qi2017pointnet}} 
        &\quad
        &\multicolumn{4}{c}{PointNet++ \cite{qi2017pointnet++}}       \\
        \cline{2-5} \cline{7-10}
        
        &\makecell[c]{CA}
        &\makecell[c]{IA}
        &\makecell[c]{$\tau$=0.1}
        &\makecell[c]{$\tau$=0.5}
        &\quad
        &\makecell[c]{CA}
        &\makecell[c]{IA}
        &\makecell[c]{$\tau$=0.1}
        &\makecell[c]{$\tau$=0.5}         \\
        \cline{2-5} \cline{7-10}

        Without DLF
        &\makecell[c]{87.1}
        &\makecell[c]{90.6}
        &\makecell[c]{64.3}
        &\makecell[c]{{2.5}}
        &\quad
        &\makecell[c]{{90.5}}
        &\makecell[c]{{92.5}}
        &\makecell[c]{67.1}
        &\makecell[c]{2.6}         \\

        \rowcolor{gray!20} 
        With DLF
        &\makecell[c]{{88.5}}
        &\makecell[c]{{91.7}}
        &\makecell[c]{{70.1}}
        &\makecell[c]{{12.4}}
        &\quad
        &\makecell[c]{{91.5}}
        &\makecell[c]{{93.8}}
        &\makecell[c]{{72.5}}
        &\makecell[c]{{14.3}}   \\

        \Xhline{1pt}
	\end{tabular}
	%\end{center}
 }
 \vspace{-5pt}
	\caption{The classification results on ModelNet40 \cite{wu20153d}. `CA`/'IA' means class accuracy/instance accuracy.}
	\label{tab542}

\end{table}

\vspace{-15pt}
\subsection{Ablation Study} 

\begin{table}[h]
        \scriptsize
	%\begin{center}
  \resizebox{0.475\textwidth}{!}{
	\begin{tabular}{p{1.3cm}p{0.8cm}p{0.8cm}p{0.8cm}p{0.005cm}p{1.0cm}p{1.0cm}p{1.0cm}p{1.0cm}}	
        \Xhline{1pt}
  
        \multirow{2}{*}{Methods}
        &\multirow{2}{*}{$\#$Params}
        &\makecell[c]{\multirow{2}{*}{NDS}} 
        &\makecell[c]{\multirow{2}{*}{mAP}} 
        &\quad
        &\multicolumn{4}{c}{Robustness (mAP)}  \\
         \cline{6-9}

        &
        &
        &
        &\quad
        &\makecell[c]{$\tau$=0.05}
        &\makecell[c]{$\bm{T}$=0.5}
        &\makecell[c]{$S$=0.95}
        &\makecell[c]{$R$=0.05}\\
       \hline

        Baseline
        &\makecell[c]{11.1M}
        &\makecell[c]{71.2}
        &\makecell[c]{68.0}
        &\quad
        &\makecell[c]{59.8}
        &\makecell[c]{46.9}
        &\makecell[c]{34.1}
        &\makecell[c]{37.7}\\

        RSDNet*
        &\makecell[c]{15.8M}
        &\makecell[c]{71.1}
        &\makecell[c]{67.9}
        &\quad
        &\makecell[c]{59.4}
        &\makecell[c]{45.3}
        &\makecell[c]{33.2}
        &\makecell[c]{36.4}\\

        RSDNet$_{SGCL}$
        &\makecell[c]{15.8M}
        &\makecell[c]{71.5}
        &\makecell[c]{68.6}
        &\quad
        &\makecell[c]{62.9}
        &\makecell[c]{50.4}
        &\makecell[c]{38.1}
        &\makecell[c]{38.9}\\

        \rowcolor{gray!20} 
        RSDNet
        &\makecell[c]{16.5M}
        &\makecell[c]{71.9}
        &\makecell[c]{68.9}
        &\quad
        &\makecell[c]{64.0}
        &\makecell[c]{51.6}
        &\makecell[c]{39.3}
        &\makecell[c]{39.9}\\

        \Xhline{1pt}
        
	\end{tabular}
	%\end{center}
 }
 \vspace{-5pt}
	\caption{Ablation study of the denoising learning in RSDNet on nuScenes. The baseline (Fully Sparse Pipeline, FSP) means removing DUNet from RSDNet. Meanwhile, RSDNet* denotes removing the denoising process from RSDNet, but retaining the DUNet targeting ${\bm{x_0^{lat}}}$. RSDNet$_{SGCL}$ indicates removing SGCL from RSDNet. This results demonstrate the importance of the denoising learning for RSDNet.}
	\label{tab551}

\end{table}

\noindent\textbf{The Denoising Learning.} We first conduct the ablation study for the denoising learning in RSDNet. Tab.~\ref{tab551} shows the results on nuScenes. RSDNet* exhibits  a significant drop across all evaluation metrics. Removing the conditional denoising learning under the task-specific knowledge guidance hinders learning robust and generalizable representations for RSDNet, reducing the object detection performance in unseen scenes, especially for noise-sensitive small objects. This also suggests that simply increasing model capacity cannot guarantee the better detection performance and may even lead to overfitting (RSDNet* is lower than  baseline).

\begin{figure}[htp]
	\centering
\includegraphics[width=0.47\textwidth]
 {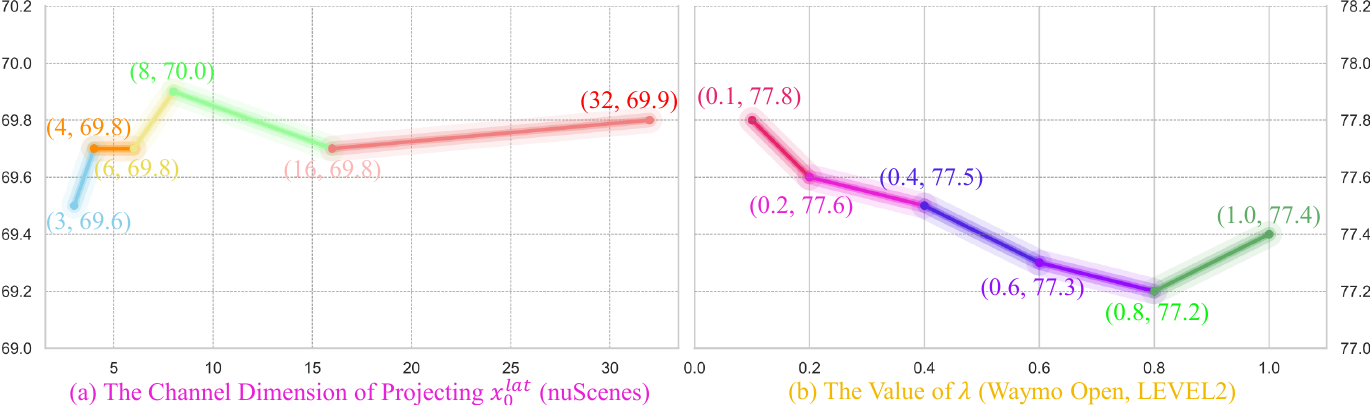}
 \vspace{-15pt}
\caption{ The y-axis represents mAP. (a) The ablation study of the  channel dimension for projecting $\bm{x_0^{lat}}$ on nuScenes. (b) The ablation study for the value of $\lambda$ on Waymo Open.}
	\label{fig7}
    \vspace{-6mm} 
\end{figure}

\noindent\textbf{The Channel Dimension of Projecting $\bm{x_0^{lat}}$.} As mentioned in Sec.~4.1, DUNet should expect to be lightweight. Therefore, we explore the impact of projecting $\bm{x_0^{lat}}$ into different latent spaces for RSDNet. Fig.~\ref{fig7}(a) shows that projecting $\bm{x_0^{lat}}$ into 8 channels yields the best trade-off between performance and efficiency. Meanwhile, further increasing the channel dimensions leads to saturated or even degraded performance. We believe that the larger channel dimension may introduce more unreasonable perturbations,  impairing the effectiveness of the denoising learning for RSDNet.

\noindent\textbf{The Value of $\lambda$.} Furthermore, we also conduct the ablation study for the value of $\lambda$, as illustrated in Fig.~\ref{fig7}(b). We can observe that when $\lambda$ becomes large, the performance of RSDNet drops significantly. This further validates that excessive perturbations can impair the denoising learning effect for RSDNet, hindering effectively understanding the scene context under the multi-type and multi-level perturbations.

\section{Conclusion} In this paper, we  revealed the source of robustness and reformulated the noising and denoising rules. Building upon these core insights, a Detachable Latent Framework of DDPMs was designed to overcome multi-step inference and model multiple perturbations. Furthermore, based on this, we proposed a robust single-stage fully sparse object detection network, exhibiting superior robustness and performance. Overall, we provided new knowledge  for applying DDPMs, hoping to inspire further extensions in 3D tasks.

\bibliography{aaai2026}

\newpage

\setcounter{equation}{0}
\setcounter{section}{0}
\setcounter{figure}{0}
\setcounter{table}{0}

Due to the space limitation of the main text, we include additional experiments, derivations, implementations, and discussions in the supplementary material. We first conduct additional ablation (Sec.~1). Then, we logically derive DDPMs from a modeling perspective, explaining some issues about the distribution matching and the sample fitting  (Sec.~2). Next, the implementation details (Sec.~3) of our method are presented. Finally, we discuss the limitations of DLF (Sec.~4) and visualize additional results (Sec.~5).

\section{Additional Ablation Study}

\subsection{Training Objective}

As described in Sec.~4.1 of the main text, RSDNet can actually adopt four types of training objectives:  $\bm{x_0^{lat}}$, $g_t^*(\bm{x_0^{lat}})$, $\bm{\epsilon_{t-1}}$ and $h_t^*(\bm{\epsilon_{t-1}})$. Tab.~\ref{supp_tab111} presents the result. The training objectives $\bm{x_0}$ and $g_t^*(\bm{x_0})$ exhibit the significant drop in the noise robustness and detection performance, which is consistent with the findings of some previous works \cite{ho2020denoising, qu2025end}. Intuitively, since $\bm{\epsilon_{t-1}}$ and $h_t^*(\bm{\epsilon_{t-1}})$ are closer to the predefined prior distribution, their structures are simpler compared to those of $\bm{x_0}$ and $g_t^*(\bm{x_0})$  from the unknown data distribution. Therefore, this makes the easier for the network to fit. In fact, this may also mean that the noise targets contribute more to the noise robustness than the noise samples.

\vspace{-5pt}
\begin{table}[h]
        \scriptsize
	%\begin{center}
  \resizebox{0.475\textwidth}{!}{
	\begin{tabular}{p{1.3cm}p{0.8cm}p{0.8cm}p{0.005cm}p{1.0cm}p{1.0cm}p{1.0cm}p{1.0cm}}	
        \Xhline{1pt}
  
        \multirow{2}{*}{Objective}
        &\makecell[c]{\multirow{2}{*}{NDS}} 
        &\makecell[c]{\multirow{2}{*}{mAP}} 
        &\quad
        &\multicolumn{4}{c}{Robustness (mAP)}  \\
         \cline{5-8}

        &
        &
        &\quad
        &\makecell[c]{$\tau$=0.05}
        &\makecell[c]{$\bm{T}$=0.5}
        &\makecell[c]{$S$=0.95}
        &\makecell[c]{$R$=0.05}\\
       \hline

        $\bm{x_0}$
        &\makecell[c]{71.2}
        &\makecell[c]{68.5}
        &\quad
        &\makecell[c]{62.8}
        &\makecell[c]{45.5}
        &\makecell[c]{34.9}
        &\makecell[c]{38.0}\\

        $g_t^*(\bm{x_0})$
        &\makecell[c]{71.4}
        &\makecell[c]{68.4}
        &\quad
        &\makecell[c]{62.3}
        &\makecell[c]{49.1}
        &\makecell[c]{37.2}
        &\makecell[c]{38.3}\\

        $\bm{\epsilon}$
        &\makecell[c]{71.8}
        &\makecell[c]{68.8}
        &\quad
        &\makecell[c]{64.5}
        &\makecell[c]{47.8}
        &\makecell[c]{36.4}
        &\makecell[c]{38.5}\\

        \rowcolor{gray!20} 
        $h_t^*(\bm{\epsilon})$
        &\makecell[c]{71.9}
        &\makecell[c]{68.9}
        &\quad
        &\makecell[c]{64.0}
        &\makecell[c]{51.6}
        &\makecell[c]{39.3}
        &\makecell[c]{39.9}\\

        \Xhline{1pt}
        
	\end{tabular}
	%\end{center}
 }
 \vspace{-5pt}
	\caption{The ablation study of the training objective in RSDNet on nuScenes. The training objectives $\bm{x_0}$ and $g_t^*(\bm{x_0})$ for RSDNet exhibit the significant degradation in the noise robustness and the detection performance. Meanwhile, fitting $h_t^*(\bm{\epsilon_{t-1}})$ in RSDNet shows the better result.}
	\label{supp_tab111}

\end{table}

\vspace{-15pt}

\subsection{Time Embedding}

Since DUNet consists of two sub-networks with identical architectures, we further investigate the impact of the $t$ embeddings within 3DDU and 2DDU for RSDNet. As shown in Tab.~\ref{supp_tab121}, the hybrid  
$t$ embeddings yield better results. We believe that the hybrid $t$ embeddings allow RSDNet to learn the richer denoising path,  can perceive a wider range of reasonable multi-type and multi-level perturbations in one epoch, thus improving the robustness and generalization for the backbone features.

\begin{table}[h]
        \scriptsize
	%\begin{center}
  \resizebox{0.475\textwidth}{!}{
	\begin{tabular}{p{1.3cm}p{0.8cm}p{0.8cm}p{0.005cm}p{1.0cm}p{1.0cm}p{1.0cm}p{1.0cm}}	
        \Xhline{1pt}
  
        \multirow{2}{*}{Embedding}
        &\makecell[c]{\multirow{2}{*}{NDS}} 
        &\makecell[c]{\multirow{2}{*}{mAP}} 
        &\quad
        &\multicolumn{4}{c}{Robustness (mAP)}  \\
         \cline{5-8}

        &
        &
        &\quad
        &\makecell[c]{$\tau$=0.05}
        &\makecell[c]{$\bm{T}$=0.5}
        &\makecell[c]{$S$=0.95}
        &\makecell[c]{$R$=0.05}\\
       \hline

        $t^{-}$
        &\makecell[c]{71.2}
        &\makecell[c]{68.5}
        &\quad
        &\makecell[c]{62.8}
        &\makecell[c]{45.5}
        &\makecell[c]{34.9}
        &\makecell[c]{38.0}\\

        $t^{+}$
        &\makecell[c]{71.7}
        &\makecell[c]{68.6}
        &\quad
        &\makecell[c]{62.3}
        &\makecell[c]{48.1}
        &\makecell[c]{37.2}
        &\makecell[c]{38.3}\\

        \rowcolor{gray!20} 
        $t^{*}$
        &\makecell[c]{71.9}
        &\makecell[c]{68.9}
        &\quad
        &\makecell[c]{64.0}
        &\makecell[c]{51.6}
        &\makecell[c]{39.3}
        &\makecell[c]{39.9}\\

        \Xhline{1pt}
        
	\end{tabular}
	%\end{center}
 }
 \vspace{-5pt}
	\caption{The ablation study of the time embedding in RSDNet on nuScenes. $t^{-}$ means the removal of time embedings in RSDNet. $t^{+}$ indicates  the same time embeddings. Meanwhile, $t^{*}$ presents  the hybrid time embeddings. }
	\label{supp_tab121}
\vspace{-3mm}
\end{table}

\subsection{Noise Construction Module}

NCM constructs the multi-type and multi-level noise samples and targets. In practice, we can still choose to model only Gaussian distribution without NCM in RSDNet. As shown in Tab.~\ref{supp_tab131}, RSDNet modeled only Gaussian distribution struggles to remain robust for coordinate translation, scaling, and rotation. Meanwhile, while multi-type perturbations leads to the training difficulty and may slightly degrade the detection performance, we believe that the gain in robustness outweighs the loss in detection accuracy.

\vspace{-5pt}
\begin{table}[h]
        \scriptsize
	%\begin{center}
  \resizebox{0.475\textwidth}{!}{
	\begin{tabular}{p{1.8cm}p{0.8cm}p{0.8cm}p{0.005cm}p{1.0cm}p{1.0cm}p{1.0cm}p{1.0cm}}	
        \Xhline{1pt}
  
        \multirow{2}{*}{Method}
        &\makecell[c]{\multirow{2}{*}{NDS}} 
        &\makecell[c]{\multirow{2}{*}{mAP}} 
        &\quad
        &\multicolumn{4}{c}{Robustness (mAP)}  \\
         \cline{5-8}

        &
        &
        &\quad
        &\makecell[c]{$\tau$=0.05}
        &\makecell[c]{$\bm{T}$=0.5}
        &\makecell[c]{$S$=0.95}
        &\makecell[c]{$R$=0.05}\\
       \hline

        Without NCM
        &\makecell[c]{71.9}
        &\makecell[c]{69.0}
        &\quad
        &\makecell[c]{64.9}
        &\makecell[c]{47.3}
        &\makecell[c]{37.0}
        &\makecell[c]{37.6}\\

        \rowcolor{gray!20} 
        With NCM
        &\makecell[c]{71.9}
        &\makecell[c]{68.9}
        &\quad
        &\makecell[c]{64.0}
        &\makecell[c]{51.6}
        &\makecell[c]{39.3}
        &\makecell[c]{39.9}\\

        \Xhline{1pt}
        
	\end{tabular}
	%\end{center}
 }
 \vspace{-5pt}
	\caption{The ablation study of NCM in RSDNet on nuScenes. Constructing multi-type and multi-level noise samples and targets via NCM shows the better trade-off between the noise robustness and the detection performance. }
	\label{supp_tab131}

\end{table}

\begin{figure*}[htp]
	\centering
\includegraphics[width=\textwidth]
 {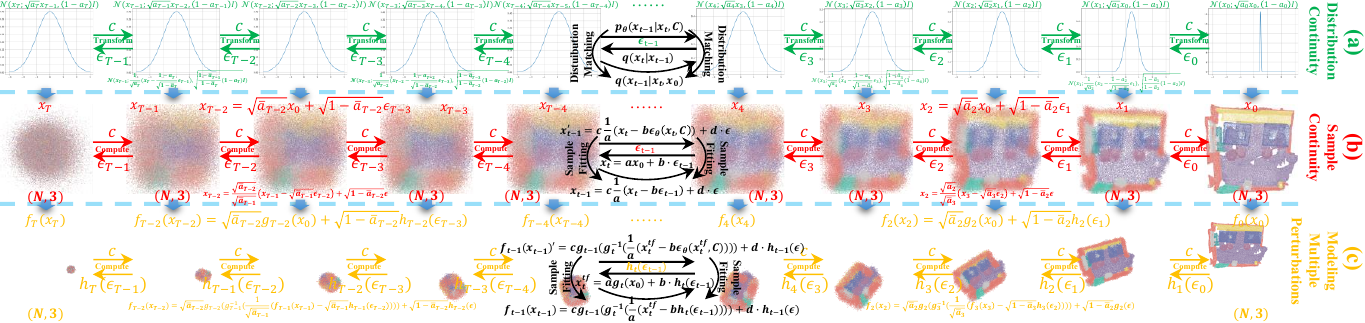}
\vspace{-18pt} 
\caption{Visualization of distribution matching, sample fitting, and modeling multiple perturbations. (a) The distribution matching requires each step to follow continuous and correlated distribution transformations, forming a complex formula chain \cite{ho2020denoising}. (b) The sample fitting focuses on effectively estimating the next-step sample, simplifying the construction conception \cite{bansal2023cold}. (c) The multi-type and multi-level noise samples and targets can be constructed using affine transformations. For example, $S_t(\bm{x_t}$$)=$$\sqrt{\overline{\alpha}_t}S_t(\bm{x_0}) +  \sqrt{1-\overline{\alpha}_t}S_t(\bm{\epsilon_{t-1}})$, this implements an $S_t$-fold scaling of $\bm{x_t}$.}
	\label{supp_fig2}
\end{figure*}

\vspace{-15pt}
\section{Formula Derivation of DDPMs}

In this section, we provide the theoretical support for the distribution matching, the sample fitting and the modeling multiple perturbations. To the easier understanding, this focuses on the logical and intuitive derivation, omitting some derivation details that we consider unnecessary. For example, we believe that the derivation of the
\textbf{E}vidence \textbf{L}ower \textbf{BO}und (ELBO) can be skipped without affecting the full understanding of the DDPM derivation. Interested readers can refer to the supplementary material of \cite{qu2024conditional} for the detail derivation of ELBO under specific conditions. 

Given an observed sample $\bm{c} \sim P_{sample}$, a fitting target $\bm{x_0} \sim P_{target}$, and a latent variable $\bm{x_T} \sim P_{noise}$, DDPMs achieve \textbf{\textit{the distribution transformation process between $P_{target}$ and $P_{noise}$}} via: a predefined diffusion process $q$ that gradually adds noise to $\bm{x_0}$ until $\bm{x_0}$ degrades into $\bm{x_T}$, and a trainable generation process $p_\theta$ that slowly removes perturbation from $\bm{x_T}$ until $\bm{x_T}$ recovers $\bm{x_0}$ conditioned on $\bm{c}$.

\subsection{Affine Transformation}

We can apply a linear transformation coefficient $a$ or a translation vector $\bm{b}$ to the variable $\bm{x}$ to produce an affine transformation $f^*(\bm{x})$:

\vspace{-10pt}
\begin{equation}
\begin{split}
\label{supp_f211}
	f^*(\bm{x}) = a\bm{x} + \bm{b}.
\end{split}
\end{equation}

In 3D tasks, most non-trained data augmentation methods, such as translation, scaling and rotation, can be implemented through Eq.~\ref{supp_f211}.

\vspace{-10pt}
\begin{equation}
\begin{split}
\label{supp_f212}
    Translation(\bm{x})=\bm{x}-\bm{T},\\
	Scaling(\bm{x})=S\bm{x}, \quad \quad\\ 
    Rotation(\bm{x})=R \cdot \bm{x}, \quad
\end{split}
\end{equation}
where $\bm{T}$ represents a translation vector, $S$ indicates a scaling factor and $R$ denotes a rotation matrix.

\textbf{\textit{Lemma: The composition function of invertible affine transformation functions is still an invertible function.}}

\subsection{Noising}

In this section, we first derive the traditional predefined diffusion process (the original noising mechanism), then provide a way for noising multiple perturbations.

The diffusion process $q$ is more crucial, as this defines the modeled distribution type of DDPMs, constructing the training samples. This typically is modeled as an independent Gaussian distribution and a Markov process \cite{ho2020denoising}. 

\textbf{The Original Noising.} We can logically and intuitively understand this process. According to Fig.~\ref{supp_fig2}(a), sampling $\bm{x_1}$ is conditioned on $\bm{x_0}$ as a prior, $\bm{x_1} \sim q(\bm{x_1}|\bm{x_0})$. Similarly, sampling $\bm{x_2}$ is conditioned on $\bm{x_1}$ and $\bm{x_0}$ as priors, $\bm{x_2} \sim q(\bm{x_2}|\bm{x_1}, \bm{x_0})$. In this way, sampling $\bm{x_T} \sim q(\bm{x_T}|\bm{x_{0:T-1}})$.  Due to the independent and
identically distributed (i.i.d.) and Markov properties, \textbf{\textit{this forward distribution transformation process}} can be described as:

\begin{equation}
\begin{split}
	\label{supp_f221}
\textcolor{blue}{i.i.d.:} \quad\quad\quad\quad\quad\quad\quad\quad\quad\quad\quad\quad\quad\quad\quad\quad\quad\quad\quad\quad\quad\quad\quad\quad\quad\quad\quad\quad\quad\quad\quad\quad\;\;\\ 
q(\bm{x_1}|\textcolor{red}{\bm{x_0}})q(\bm{x_2}|\textcolor{red}{\bm{x_1}},\textcolor{red}{\bm{x_0}})...q(\bm{x_T}|\textcolor{red}{\bm{x_{0:T-1}}})\quad\quad\quad\quad\quad\quad\quad\quad\quad\quad\quad\quad \\
\textcolor{blue}{Markov:} \quad\quad\quad\quad\quad\quad\quad\quad\quad\quad\quad\quad\quad\quad\quad\quad\quad\quad\quad\quad\quad\quad\quad\quad\quad\quad\quad\quad\quad\quad\;\;\;\\
=q(\bm{x_1}|\textcolor{red}{\bm{x_0}})q(\bm{x_2}|\textcolor{red}{\bm{x_1}})...q(\bm{x_T}|\textcolor{red}{\bm{x_{T-1}}})\quad\quad\quad\quad\quad\quad\quad\quad\quad\quad\quad\quad\quad\quad\;\\
= \prod_{t=1}^{T}q(\bm{x_t}|\bm{x_{t-1}}).\quad\quad\quad\quad\quad\quad\quad\quad\quad\quad\quad\quad\quad\quad\quad\quad\quad\quad\quad\quad\quad\;\;
\end{split}
\end{equation}

Meanwhile, to construct the sample $\bm{x_t}$,  a reparameterization trick (inverse transform sampling) to sample $\bm{x_t}$ from a specified mean and variance: $\bm{x_t}=\bm{\mu_t}+{\sigma_t}\bm{\epsilon_{t-1}}$,  $\bm{\epsilon_{t-1}} \sim \mathcal{N}(\bm{\epsilon_{t-1}}; \bm{0}, \bm{I})$. This sampling $\bm{x_t}$ follows the Gaussian distribution with mean $\bm{\mu_t}$ and variance $\sigma_t^2\bm{I}$ (the properties of random variable). Then, a variance schedule $\beta \in [0.0001, 0.02]$ at increases with $t$ is first predefined. Therefore, $\bm{x_t} = [\sqrt{1-\beta_t}\bm{x_{t-1}}]_{\bm{\mu}} + [\sqrt{\beta_t}]_{{\sigma}} \cdot \bm{\epsilon_{t-1}}$. Meanwhile, we can further simplify to compute   $\bm{x_t}$ by setting $\alpha_t = 1 - \beta_t$, and $\overline{\alpha}_t = \prod_{t=1}^{T}\alpha_t$:

\begin{equation*}
    \begin{split}
        \bm{x_t}=[\sqrt{1-\beta_t}\bm{x_{t-1}}]_{\bm{\mu}} + [\sqrt{\beta_t}]_{{\sigma}} \cdot \bm{\epsilon_{t-1}}\quad\quad\quad\quad\quad\quad\;\;\;\;\;\;\\
=[\sqrt{\alpha_t}\bm{x_{t-1}}]_{\bm{\mu}} + [\sqrt{1-\alpha_t}]_{{\sigma}} \cdot \bm{\epsilon_{t-1}}\quad\quad\quad\quad\quad\quad\quad\;\;\;\\ 
    \end{split}
\end{equation*}

\begin{equation}
    \begin{split}
    \label{supp_f222} 
=\sqrt{\alpha_t}(\sqrt{\alpha_{t-1}}\bm{x_{t-2}} + \sqrt{1-\alpha_{t-1}}\bm{\epsilon_{t-1}}) + \sqrt{1-\alpha_t}\bm{\epsilon_t}\;\\
=\sqrt{\alpha_t\alpha_{t-1}}\bm{x_{t-2}}+\textcolor{blue}{\sqrt{\alpha_t-\alpha_t\alpha_{t-1}}\bm{\epsilon_{t-1}}+\sqrt{1-\alpha_t}\bm{\epsilon_t}}\;\\
\textcolor{blue}{Gaussian \; Variable \;Additivity:} \quad\quad\quad\quad\quad\quad\quad\quad\quad\quad\;\\
\textcolor{blue}{\Rightarrow \sqrt{a_t-a_ta_{t1}}\bm{\epsilon_{t-1}} \sim \mathcal{N}(0,(a_t-a_ta_{t-1}))}\quad\quad\quad\quad\quad\;\;\;\\
\textcolor{blue}{\Rightarrow \sqrt{1-a_t}\bm{\epsilon_t} \sim \mathcal{N}(0,(1-a_t)))}\quad\quad\quad\quad\quad\quad\quad\quad\quad\quad\;\;\\
\textcolor{blue}{\Rightarrow \mathcal{N}(0,(a_t-a_ta_{t-1})+(1-a_t))}\quad\quad\quad\quad\quad\quad\quad\quad\;\;\;\;\;\;\\
=\sqrt{\alpha_t\alpha_{t-1}}\bm{x_{t-2}}+\textcolor{blue}{\sqrt{1-\alpha_t\alpha_{t-1}}\bm{\epsilon}}\quad\quad\quad\quad\quad\quad\quad\;\;\;\\
...\quad\quad\quad\quad\quad\quad\quad\quad\quad\quad\quad\quad\quad\quad\quad\quad\quad\quad\quad\quad\quad\;\;\\
=[\sqrt{\overline{\alpha}_t}\bm{x_0}]_{\bm{\mu}} + [\sqrt{1-\overline{\alpha}_t}]_{{\sigma}} \cdot \bm{\epsilon}. \quad\quad\quad\quad\quad\quad\quad\quad\quad\;\;\;
    \end{split}
\end{equation}

Therefore, in the predefined diffusion process, we actually obtain the means and variances of $q(\bm{x_t}|\bm{x_{t-1}})$ and $q(\bm{x_t}|\bm{x_0})$. Meanwhile, according to Eq.~\ref{supp_f222}, we can observe that \textbf{\textit{the sample construction in DDPMs actually relies on $q(\bm{x_t}|\bm{x_0})$ rather than $q(\bm{x_t}|\bm{x_{t-1}})$}}.

\textbf{Multiple Perturbation Noising.} Next, based on $q(\bm{x_t}|\bm{x_0})$ in Eq.~\ref{supp_f222}, we can conveniently apply affine transformations to $\bm{x_0}$ and $\bm{\epsilon}$ to construct  $\bm{x_t}$ with multiple perturbations (see Fig.\ref{supp_fig2}(c)):

\begin{equation}
    \begin{split}
    \label{supp_f223}
    \bm{x_t}
=\sqrt{\overline{\alpha}_t}\bm{x_0} + \sqrt{1-\overline{\alpha}_t} \cdot \bm{\epsilon_{t-1}}, \quad\quad\quad\quad\quad\quad\quad\quad\quad\;\;\;\\
\textcolor{blue}{\Rightarrow let \;a=\sqrt{\overline{\alpha}_t}, b=\sqrt{1-\overline{\alpha}_t}}\quad\quad\quad\quad\quad\quad\quad\quad\quad\quad\quad\quad\quad\quad\;\\
\bm{x_t}=a\bm{x_0} + b \cdot \bm{\epsilon_{t-1}},\quad\quad\quad\quad\quad\quad\quad\quad\quad\quad\quad\quad\quad\;\;\;\;\\
\textcolor{blue}{\Rightarrow affine \; transformation:} \quad\quad\quad\quad\quad\quad\quad\quad\quad\quad\quad\quad\quad\quad\\
\textcolor{red}{\bm{x_t^{tf}}=f_t(\bm{x_t})=ag_t(\bm{x_0}) + b \cdot h_t(\bm{\epsilon_{t-1}})}, \quad\quad\quad\quad\quad\quad\;\;\\
    \end{split}
\end{equation}
where $g_t(\cdot)$ and $h_t(\cdot)$ indicate invertible affine transformation functions with intensity varying base on $t$.  $f_t(\cdot)$is a invertible affine transformation determined by $g_t(\cdot)$ and $h_t(\cdot)$. 

Therefore, we obtain the multiple perturbation noising formulation as shown in Eq.~5 of the main text. \textbf{\textit{This reformulates the noising mechanism as a more general formulation to construct multi-type perturbations in DDPMs.}} When $g_t(\bm{x_0})=1\cdot\bm{x_0}$ and $h_t(\bm{\epsilon})=1\cdot\bm{\epsilon}$, Eq.~\ref{supp_f223} transforms to the original noising formulation in Eq.~\ref{supp_f222}.

\textbf{\textit{We cannot directly apply affine transformations to $q(\bm{x_t}|\bm{x_{t-1}})$, \ie, $ f_t(\bm{x_t})=\sqrt{1-\beta_t}g_t(\bm{x_{t-1}}) + \sqrt{\beta_t}h_t(\bm{\epsilon_{t-1}})$.}} This is because $\bm{x_t}$ is determined by $\bm{x_{t-1}}$, which in turn is determined by $\bm{x_{t-2}}$, and so on. This creates a complex formulation chain (as discussed in Sec.~3.3 of the main text). The relationships between samples ($\bm{x_0}...\bm{x_t}$) are highly entangled.  The becomes extremely difficult to convert from $q(\bm{x_t}|\bm{x_{t-1}})$ to $q(\bm{x_t}|\bm{x_0})$ in Eq.~\ref{supp_f222}, making the derivation of $q(\bm{x_{t-1}}|\bm{x_t},\bm{x_0})$ particularly challenging.

\subsection{Denoising}

In this section, we first derive the ground truth of DDPMs (the original denoising mechanism), then presenting the denoising process under the modeling multiple perturbations.

In the forward distribution transition process, all variables ($\bm{x_0}$ and $\bm{\epsilon}$) are known. Therefore, we can inverse this process to obtain the inverse process (the true posterior of the generative process, the Ground Truth), $q(\bm{x_{t-1}}|\bm{x_{t}}, \bm{x_0})$. \textbf{\textit{This inverse process must be calculated based on the prior $\bm{x_0}$, since this predefined diffusion process must be performed first (from $\bm{x_0}$ to $\bm{x_T}$).}} To match the forward distribution transition process, this inverse process is also modeled as an independent Gaussian distribution and a Markov chain. 

\textbf{The Original Denoising.} This can be directly derived using the inverse probability rule (Bayes' theorem) for $q(\bm{x_t}|\bm{x_{t-1}})$ in Eq.~\ref{supp_f221}:

\begin{equation}
\begin{split}
\label{supp_f231}
q(\bm{x_t}|\bm{x_{t-1}},\textcolor{red}{\bm{x_0}})=\frac{\textcolor{red}{q(\bm{x_{t-1}}|\bm{x_t},\bm{x_0})}q(\bm{x_t}|\textcolor{red}{\bm{x_0})}}{q(\bm{x_{t-1}}|\textcolor{red}{\bm{x_0})}}\quad\quad\quad\\
\textcolor{red}{q(\bm{x_{t-1}}|\bm{x_t},\bm{x_0})}=\frac{q(\bm{x_t}|\bm{x_{t-1}},\bm{x_0})q(\bm{x_{t-1}}|\bm{x_0})}{q(\bm{x_t}|\bm{x_0})}\quad\quad\\
\textcolor{blue}{Markov:} \quad\quad\quad\quad\quad\quad\quad\quad\quad\quad\quad\quad\quad\quad\quad\quad\quad\quad\quad\\
=\frac{\textcolor{black}{q(\bm{x_t}|\bm{x_{t-1}})}q(\bm{x_{t-1}}|\bm{x_0})}{q(\bm{x_t}|\bm{x_0})},\quad\quad\;\;\;\;
\end{split}
\end{equation}
where $q(\bm{x_t}|\bm{x_{t-1}}) = \mathcal{N}(\bm{x_t};\sqrt{\alpha_t}\bm{x_{t-1}},(1-\alpha_t)\bm{I})$, $q(\bm{x_{t-1}}|\bm{x_0}) = \mathcal{N}(\bm{x_{t-1}};\sqrt{\overline{\alpha}_{t-1}}\bm{x_0},(1-\overline{\alpha}_{t-1})\bm{I})$, and $q(\bm{x_t}|\bm{x_0}) = \mathcal{N}(\bm{x_t};\sqrt{\overline{\alpha}_t}\bm{x_0},(1-\overline{\alpha}_t)\bm{I})$. These distributions are known in the predefined diffusion process derivation. Subsequently, by substituting $q(\bm{x_t}|\bm{x_{t-1}})$, $q(\bm{x_{t-1}}|\bm{x_0})$ and $q(\bm{x_t}|\bm{x_0})$ into Eq \ref{supp_f231}, the mean $\bm{{\mu_t}}$ and the variance ${\sigma_t^2}\bm{I}$ of $q(\bm{x_t}|\bm{x_{t-1}},\bm{x_0})=\mathcal{N}(\bm{x_t};\bm{\mu_t}, \sigma_t^2\bm{I})$ can be obtain:

\begin{equation}
\begin{split}
\label{supp_f232}
\bm{\mu_t}=\frac{\sqrt{\alpha_t}(1-\overline{\alpha}_{t-1})}{1-\overline{\alpha_t}}\bm{x_t}+\frac{\sqrt{\overline{\alpha}_{t-1}}(1-\alpha_t)}{1-\overline{\alpha}_t}\bm{x_0},\\
{\sigma_t^2}=\frac{1-\overline{\alpha}_{t-1}}{1-\overline{\alpha}_t}(1-\alpha_t)\bm{I}.\quad\quad\quad\quad\quad\quad\quad\quad\;
\end{split}
\end{equation}

Meanwhile, in Eq.~\ref{supp_f232}, we can observe that the variance ${\sigma_t^2}\bm{I}$ is a constant term (some works also set the variance as a fitting term \cite{nichol2021improved}). 

Next, due to the better performance observed in experiment \cite{ho2020denoising}, $\bm{x_0}$ is considered to be replaced by $\bm{\epsilon}$, \ie, $\bm{x_0}=\frac{\bm{x_t}-\sqrt{1-\overline{\alpha}_t}\bm{\epsilon_{t-1}}}{\sqrt{\overline{\alpha}_t}}$:

\begin{equation}
\begin{split}
\label{supp_f233}
\bm{\mu_t}=\frac{1}{\sqrt{\alpha_t}}(\bm{x_t}-\frac{1-\alpha_t}{\sqrt{1-\overline{\alpha}_t}}\bm{\epsilon}).
\end{split}
\end{equation}

Notably, the initial value $\bm{x_T}$ of $\bm{x_t}$ can be directly sampled from a prior distribution $P_{noise}$ (the Gaussian distribution, $\bm{x_T} 
\sim \mathcal{N}(\bm{x_T};\bm{0},\bm{I})$) during inference. \textbf{\textit{Therefore, the only unknown term is $\bm{\epsilon}$ in Eq.~\ref{supp_f233}.}} 

Subsequently, we can sample from the specified Gaussian distribution $q(\bm{x_{t-1}}|\bm{x_t},\bm{x_0})$ based on the mean $\bm{\mu_t}$ and variance ${\sigma_t^2}\bm{I}$ (from $\bm{x_T}$ to $\bm{x_0}$):

\begin{equation}
\begin{split}
\label{supp_f234}
\bm{x_{t-1}}=[\frac{1}{\sqrt{\alpha_t}}(\bm{x_t}-\frac{1-\alpha_t}{\sqrt{1-\overline{\alpha}_t}}\bm{\epsilon_{t-1}})]_{\bm{\mu}} \\
+ [\sqrt{\frac{1-\overline{\alpha}_{t-1}}{1-\overline{\alpha}_t}(1-\alpha_t)}]_{\bm{\sigma}} \cdot \bm{\epsilon}. \;\;\;\; 
\end{split}
\end{equation}

Furthermore, as previously described, this inverse process corresponds to the predefined diffusion process, thus including $T$ steps. Moreover, each step must be conditioned on $\bm{x_0}$ as a prior. In fact, within the predefined diffusion process, $q(\bm{x_0}),q(\bm{x_1}|\bm{x_0}),...,q(\bm{x_T}|\bm{x_0})$ are actually known. Therefore, we can express this inverse process as a joint distribution conditioned on $\bm{x_0}$:

\begin{equation}
\begin{split}
\label{supp_f235}
q(\bm{x_0}|\bm{x_0})q(\bm{x_1}|\bm{x_0})...q(\bm{x_T}|\bm{x_0})\quad\quad\quad\quad\quad\quad\quad\quad\quad\quad\quad\\
=q(\bm{x_{0:T}}|\bm{x_0})\quad\quad\quad\quad\quad\quad\quad\quad\quad\quad\quad\quad\quad\quad\quad\;\;\\
=\frac{\textcolor{red}{q(\bm{x_{0:T}})}}{q(\bm{x_0})}\quad\quad\quad\quad\quad\quad\quad\quad\quad\quad\quad\quad\quad\quad\quad\quad\;\;\;\\
=\frac{\textcolor{red}{q(\bm{x_0}|\bm{x_{1:T}},\bm{x_0})q(\bm{x_{1:T}}, \bm{x_0})}}{q(\bm{x_0})}\quad\quad\quad\quad\quad\quad\quad\quad\;\;\;\\
=\frac{q(\bm{x_0}|\bm{x_{1:T}},\bm{x_0})q(\bm{x_1}|\bm{x_{2:T}}, \bm{x_0})...q(\bm{x_T}|\bm{x_0})\textcolor{red}{\cancel{q(\bm{x_0})}}}{\textcolor{red}{\cancel{q(\bm{x_0})}}}\\
\textcolor{blue}{Markov:}\quad\quad\quad\quad\quad\quad\quad\quad\quad\quad\quad\quad\quad\quad\quad\quad\quad\quad\;\;\;\\
=q(\bm{x_0}|\bm{x_{1}},\bm{x_0})q(\bm{x_1}|\bm{x_{2}}, \bm{x_0})...q(\bm{x_T}|\bm{x_0})\quad\quad\quad\;\;\;\;\;\\
\textcolor{blue}{\bm{x_T} \sim \mathcal{N}(\bm{x_T};\bm{0},\bm{I})} \quad\quad\quad\quad\quad\quad\quad\quad\quad\quad\quad\quad\quad\quad\quad\;\\
=q(\bm{x_T})\prod_{t=1}^T{q(\bm{x_{t-1}}|\bm{x_t},\bm{x_0})}.\quad\quad\quad\quad\quad\quad\quad\quad\quad\;
\end{split}
\end{equation}

We can observe that \textbf{\textit{the original DDPM formulation (noising and denoising) relies on a complex formulation chain, due to the denoising derivation, \ie, $q(\bm{x_{t-1}}|\bm{x_t},\bm{x_0})$}}.

\textbf{Multiple Perturbation Denoising.} 
In fact, according to the rule  $q(\bm{x_t}|\bm{x_0})$ in Eq.~\ref{supp_f223},  the sampling process can be written in a very straightforward manner \ie, $q(\bm{x_{t-1}}|\bm{x_0})$:

\begin{equation*}
    \begin{split}
        \bm{x_t}=[\sqrt{\overline{a}_t\bm{x_0}}]_{\bm{\mu}}+[\sqrt{1-\overline{a}_t}]_{\bm{\sigma}} \cdot \bm{\epsilon_{t-1}}\quad\quad\quad\quad\;\;\;\\
\textcolor{blue}{\Rightarrow \bm{x_0}=\frac{\bm{x_t}-\sqrt{1-\overline{\alpha}_t}\bm{\epsilon_{t-1}}}{\sqrt{\overline{\alpha}_t}}}\quad\quad\quad\quad\quad\quad\quad\quad\quad\quad\quad\quad\\
\bm{x_{t-1}}=\frac{\sqrt{\overline{a}_{t-1}}}{\sqrt{\overline{\alpha}_t}}(\bm{x_t}-\sqrt{1-\overline{\alpha}_t}\bm{\epsilon_{t-1}})+[\sqrt{1-\overline{a}_t}]_{\bm{\sigma}} \cdot \bm{\epsilon} \quad\\
\textcolor{blue}{\Rightarrow let \; c=\sqrt{\overline{\alpha}_{t-1}},d=\sqrt{1-\overline{\alpha}_{t-1}}}\quad\quad\quad\quad\quad\quad\quad\quad\quad\;\\
    \end{split}
\end{equation*}

\begin{equation}
\begin{split}
\label{supp_f236}
\textcolor{red}{\bm{x_{t-1}}=c\frac{1}{a}(\bm{x_t}-b\bm{\epsilon_{t-1}})+d \cdot \bm{\epsilon}} \quad\quad\quad\quad\quad\quad\quad
\end{split}
\end{equation}

Eq.~\ref{supp_f236} presents the denoising  under the sample fitting rule (see Fig~\ref{supp_fig2}(b)). In fact, \textbf{\textit{this remains consistent with the noising process by adopting the same form of $q(\bm{x_{t-1}}|\bm{x_0})$}}, thereby avoiding the complex derivation of $q(\bm{x_{t-1}}|\bm{x_t},\bm{x_0})$. This rule has already appeared in previous generative methods \cite{song2020denoising, bansal2023cold}, but without formally unified under a general framework.

Similarly, we can apply affine transformations to Eq.\ref{supp_f236}:

\begin{equation}
\begin{split}
\label{supp_f237}
\bm{x_{t-1}}=\frac{\sqrt{\overline{a}_{t-1}}}{\sqrt{\overline{\alpha}_t}}(\bm{x_t}-\sqrt{1-\overline{\alpha}_t}\bm{\epsilon_{t-1}})+[\sqrt{1-\overline{a}_t}]_{{\sigma}} \cdot \bm{\epsilon} \quad\quad\quad\quad\quad\;\;\\
\textcolor{blue}{\Rightarrow affine \; transformation:} \quad\quad\quad\quad\quad\quad\quad\quad\quad\quad\quad\quad\quad\quad\quad\;\\
\bm{x_{t-1}^{tf}}=\sqrt{\overline{\alpha}_{t-1}}(g_{t-1}(g^{-1}_t(\frac{1}{\sqrt{\overline{\alpha}_t}}(\bm{x_t^{tf}}-\sqrt{1-\overline{\alpha}_t}h_t(\bm{\epsilon_{t-1}})))) \quad\quad\quad\\
+\sqrt{1-\overline{\alpha}_{t-1}} \cdot h_{t-1}(\bm{\epsilon}),\quad\quad\quad\quad\quad\quad\quad\quad\quad\quad\quad\quad\quad\quad\quad\;\;\\
\textcolor{blue}{\Rightarrow let \; c=\sqrt{\overline{\alpha}_{t-1}}, d=\sqrt{1-\overline{\alpha}_{t-1}}}\quad\quad\quad\quad\quad\quad\quad\quad\quad\quad\quad\quad\quad\;\\
\textcolor{red}{\bm{x_{t-1}^{tf}}=cg_{t-1}(g_t^{-1}(\frac{1}{a}(\bm{x_t^{tf}}-bh_t(\bm{\epsilon_{t-1}}))))+d \cdot h_{t-1}(\bm{\epsilon})} \quad\quad\quad\quad\;\;
\end{split}
\end{equation}

Eq.~\ref{supp_f237} means the Ground Truth of the denoising in Eq.~5 in the main text. The generation diversity comes from $\bm{\epsilon}$.

The multiple perturbation noising and denoising mechanisms drop "the distributions notion", focusing on  the sample fitting, \ie, $\bm{x_{t-1}^{tf}{'}}\approx\bm{x_{t-1}^{tf}}$. In theory, this allows any type of perturbations to be incorporated into DDPMs, as the core of the denoising processes lies in $q(\bm{x_{t-1}}|\bm{x_0})$ rather than  $q(\bm{x_{t-1}}|\bm{x_t},\bm{x_0})$. This also unifies the formulation between the noising and denoising stages in DDPMs, providing a more intuitive understanding.

\subsection{Trainable Generation Process}

In this section, we logically present the training objectives for the distribution matching, the sample fitting, and modeling multiple perturbations.

Since the reverse process iterates from $\bm{x_T} \sim P_{noise}$ to $\bm{x_0} \sim P_{target}$ conditioned on the prior $\bm{x_0}$, we need to use a neural network with the parameter $\bm{\theta}$ to approximate each step of the inverse process,  achieving generalized sampling. 

The generation process defines the generation mode of DDPMs: unconditional generation and  conditional generation. This takes the inverse of the predefined diffusion process as the Ground Truth.

\textbf{The Training Objective of Distribution Matching.} To better fit the inverse process, each step of the generation process is characterized by i.i.d. and Markov properties, \ie, $p_\theta(\bm{x_{t-1}}|\bm{x_t},C) \approx q(\bm{x_{t-1}}|\bm{x_t},\bm{x_0})$. Therefore, the trainable generation process represents: $p_\theta(\bm{x_T})\prod_{t=1}^T{p_\theta(\bm{x_{t-1}}|\bm{x_t}, C)}$ (unconditional generation, $C=\emptyset$). Meanwhile, according to Eq.~\ref{supp_f233}, we can logically refine this distribution matching target:

\begin{equation*}
    \begin{split}
            p_\theta(\bm{x_{t-1}}|\bm{x_t},C) \approx q(\bm{x_{t-1}}|\bm{x_t},\bm{x_0}), \\
    \end{split}
\end{equation*}

\begin{equation}
\begin{split}
	\label{supp_f241}
    \Rightarrow \bm{\mu_\theta}(\bm{x_t},C)\approx \bm{\mu_t}, \quad \quad  \quad  \quad \quad \quad \; \;\; \; \; \; \\
    \Rightarrow \bm{\epsilon_\theta}(\bm{x_t},C) \approx \bm{\epsilon}.\quad \quad  \quad  \quad \quad \quad\;\;\;\; \;\;\;\;\;
\end{split}
\end{equation}

Meanwhile, due to the inverse process with a total of $T$ steps, the final training objective is:

\vspace{-5pt}
\begin{equation}
\begin{split}
\label{supp_f242}
L(\theta)=\frac{1}{T}\sum_{t=1}^{T} D_{KL}(q(\bm{x_{t-1}}|\bm{x_t},\bm{x_0})||p_\theta(\bm{x_{t-1}}|\bm{x_t},C)),\\
=\frac{1}{T}\sum_{t=1}^{T}||\bm{\mu_t} - \bm{\mu_\theta}(\bm{x_t},t,C)||_2^2,\quad\quad\quad\quad\quad\quad\;\;\;\;\;\\
 =\mathbb{E}_{\bm{\epsilon} \sim \mathcal{N}(\bm{0},\bm{I})}||\bm{\epsilon_{t-1}} - \bm{\epsilon_\theta}(\bm{x_t},t,C)||_2^2.\quad\quad\quad\quad\;\;\;\;
\end{split}
\end{equation}

This derive the training objective in Eq.~1 of the main text. 

\textbf{The Training Objective of Sample Fitting.} As shown in Eq.~\ref{supp_f231}, Eq.~\ref{supp_f232} and Eq.~\ref{supp_f233}, the distribution matching requires a complex formulation chain (deriving $q(\bm{x_{t-1}}|\bm{x_t}, \bm{x_0}$), increasing the difficulty of constructing DDPMs. In contrast, the sample fitting based on $p_\theta(\bm{x_{t-1}}|\bm{x_0})$ can simplify this construction process. Similarly, according to Eq.~\ref{supp_f237}, we further refine the objective of sample fitting in a logical manner:

\vspace{-5pt}
\begin{equation}
\begin{split}
\label{supp_f243}
\bm{x_{t-1}^{tf}{'}} \approx \bm{x_{t-1}^{tf}},\quad\quad\quad\;\\
\Rightarrow \bm{\epsilon_\theta}(\bm{x_{t}^{tf}},C) \approx h_t(\bm{\epsilon_{t-1}})
\end{split}
\end{equation}

Similarly, the final training objective for  $T$ steps is:

\vspace{-5pt}
\begin{equation}
\begin{split}
\label{supp_f244}
L(\theta)=
\frac{1}{T}\sum_{t=1}^{T}||\bm{x_{t-1}^{tf}} - 
\bm{x_{t-1}^{tf}{'}}||_2^2,\quad\quad\quad\quad\quad\quad\quad\;\;\;\;\;\\
 =\mathbb{E}_{\bm{\epsilon} \sim \mathcal{N}(\bm{0},\bm{I})}||h_t(\bm{\epsilon_{t-1}}) - \bm{\epsilon_\theta}(\bm{x_{t}^{tf}},C)||_2^2.\quad\quad\;\;\;\;
\end{split}
\end{equation}

\begin{figure*}[htp]
	\centering
\includegraphics[width=\textwidth]
 {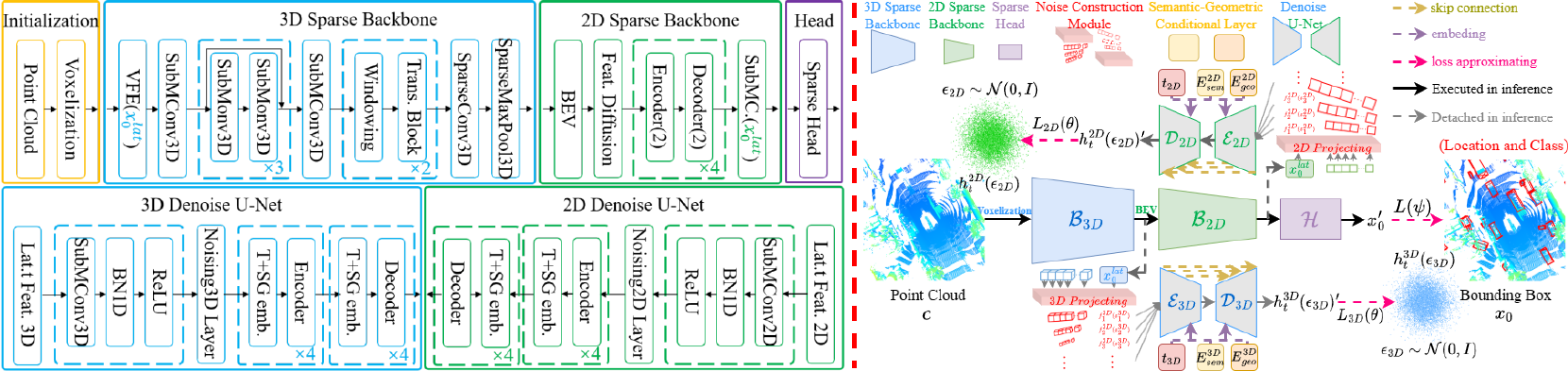}
\vspace{-10pt} 
\caption{(a) describes the six components that make up RSDNet: the Initialization Module, the 3D Sparse Backbone, the 2D Sparse Backbone, the  Head, 3DDU and 2DDU. (b) shows the overall framework of RSDNet.}
	\label{supp_fig3}
\end{figure*}

\section{Implementation}

In this section, we describe the implementation details. 

\subsection{The Config of RSDNet}

RSDNet is built upon a fully sparse detection pipeline \cite{zhang2024safdnet, liu2025fshnet}, as illustrated in Fig. \ref{supp_fig3}. The input point cloud is first voxelized by an initialization module. Then, a Voxel Feature Encoder (VFE) is used to extract graph-structured features \cite{zhang2024safdnet}. Two layers of SuMConv3D and three layers of SuMConv3D residual blocks further refine the 3D sparse representations. A two-stage linear Transformer block, similar to Swin Transformer, together with a SparseMaxPool3D layer, is adopted to expand the receptive field of the sparse features \cite{liu2025fshnet}.
After that, similar to traditional hybrid detection pipelines, the 3D sparse representation is compressed into a 2D BEV sparse representation. This is fed through a feature diffusion module \cite{zhang2024safdnet} to enhance information interaction under sparse representation.
Four sparse encoder-decoder blocks \cite{zhang2024safdnet} are employed in the BEV space to capture scene context. Finally, the 2D features are passed to the detection head to predict object locations and semantics.

In this process, 2DDU and 3DDU receive the 3D sparse features (the output $\bm{x_0^{lat}}$ from the VFE) and 2D sparse features (the output $\bm{x_0^{lat}}$ from the 2D sparse backbone), respectively, to perform multi-type and multi-level denoising learning in the 3D and 2D spaces.

The network architecture hyperparameters of RSDNet for DUnet, nuScenes and Waymo Open are shown in Tab.~\ref{supp_tab311} Tab.~\ref{supp_tab312} and Tab.~\ref{supp_tab313}, respectively.

We train RSDNet using 4 NVIDIA 3090 GPUs with batch size=8 on nuScenes and 8 NVIDIA 4090 GPUs on Waymo Open  with batch size=16, which takes 43 hours and 15 hours, respectively.

\subsection{Translation Scaling and Rotation}

To construct samples and targets with coordinate translation, scaling, and rotation in training, we apply composite \textit{\textbf{invertible affine transformations}} in the latent feature space to $\bm{x_0^{lat}}$ and $\bm{\epsilon_{t-1}}$, as defined in Eq.~6 of the main text. The translation, scaling, and rotation configurations in DUNet are detailed in Tab.~\ref{supp_tab311}.

\textbf{Translation.} We apply coordinate translation to $\bm{x_0^{lat}}$ and $\bm{\epsilon_{t-1}}$ using the following equation to construct $f_t(\bm{x_t})$:

\vspace{-5pt}
\begin{equation}
\begin{split}
\label{supp_f421}
\bm{x_t}-(\sqrt{\overline{\alpha}_t}+\sqrt{1-\overline{\alpha}_t})\bm{T_t}\quad\quad\quad\quad\quad\quad\quad\quad\\
=\sqrt{\overline{\alpha}_t}(\bm{x_0^{lat}}-\bm{T_t})+\sqrt{1-\overline{\alpha}_t}(\bm{\epsilon_{t-1}}-\bm{T_t})
\end{split}
\end{equation}

Therefore, we obtain the noise sample $\bm{x_t}-(\sqrt{\overline{\alpha}_t}+\sqrt{1-\overline{\alpha}_t})\bm{T_t}$ and  target $\bm{\epsilon_{t-1}}-\bm{T_t}$ with the coordinate offset perturbation.

\textbf{Scaling.} We apply coordinate scaling to $\bm{x_0^{lat}}$ and $\bm{\epsilon_{t-1}}$ using the following equation to construct $f_t^*(\bm{x_t})$:

\vspace{-5pt}
\begin{equation}
\begin{split}
\label{supp_f422}
S_t \cdot \bm{x_t}
=\sqrt{\overline{\alpha}_t}(S_t \cdot \bm{x_0^{lat}})+\sqrt{1-\overline{\alpha}_t}(S_t \cdot \bm{\epsilon_{t-1}})
\end{split}
\end{equation}

Therefore, we obtain the noise sample $S_t \cdot \bm{x_t}$ and  target $S_t \cdot \bm{\epsilon_{t-1}}$ with the coordinate scaling perturbation.

\textbf{Rotation.} For the rotation perturbation, we apply Givens rotations, as the latent dimension exceeds 3. Theoretically, if a 3×3 rigid rotation is applied to a point cloud coordinate,  the same transformation can always be achieved using Givens rotations. We apply different 2D Givens rigid rotation matrices respectively to the (0,1), (2,3), (4,5), and (6,7) dimensions to obtain a composite rigid Givens rotations $R_t$.  We apply coordinate rotation to $\bm{x_0^{lat}}$ and $\bm{\epsilon_{t-1}}$ using the following equation to construct $f_t^*(\bm{x_t})$:

\begin{equation}
\begin{split}
\label{supp_f423}
R_t \cdot \bm{x_t}
=\sqrt{\overline{\alpha}_t}(R_t \cdot \bm{x_0^{lat}})+\sqrt{1-\overline{\alpha}_t}(R_t \cdot \bm{\epsilon_{t-1}})
\end{split}
\end{equation}

Therefore, we obtain the noise sample $R_t \cdot \bm{x_t}$ and  target $R_t \cdot \bm{\epsilon_{t-1}}$ with the coordinate rotation perturbation.

\subsection{Integration of DLF into HEDNet and SAFDNet}

HEDNet and SADFNet are both hybrid 3D and 2D architectures. Therefore, similar to RSDNet, we integrate 3DDU and 2DDU into the 3D backbone (the output of VFE) and the 2D backbone (the output of the 2D backbone). In particular, the 2D dense features from HEDNet are converted into sparse representations before being fed into 2DDU.

\begin{figure}[htp]
	\centering
	\includegraphics[width=0.48\textwidth]
 {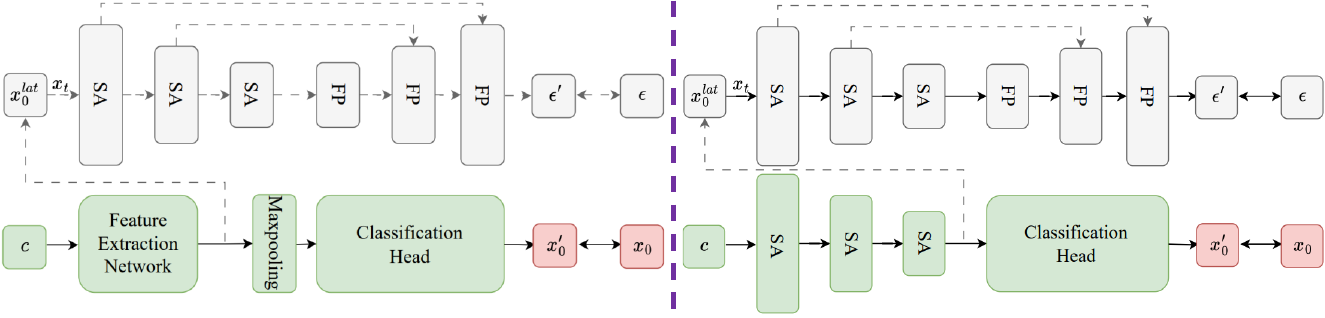}
 \vspace{-0.5cm} 
	\caption{The framework of applying DLF to PointNet and PointNet++. We use an additional PointNet++ as the denoising network.}
	\label{supp_fig4}
\end{figure}

\subsection{Integration of DLF into PointNet and PointNet++} 

Fig.~\ref{supp_fig4} illustrates the integration of DLF into the PointNet and PointNet++ frameworks. An additional PointNet++ is adopted to implement DUNet (without using NCM or SGCL). For PointNet, DUNet is applied after the feature extraction network. Meanwhile, for PointNet++, we introduce DUNet at the bottleneck stage of the U-Net architecture.

\section{Limitations}

In the main text, DLF demonstrates excellent results across various tasks, such as detection and classification. However, during the design and experimentation process, we also identified some limitations of DLF:

\begin{itemize}
    \item \textbf{Requiring More Parameters.} Although 3DDU and 2DDU contain fewer than 6M parameters and perform the denoising learning in the latent spaces, their total parameter count actually exceeds 50$\%$ of that of the baseline. Tab.~\ref{supp_tab511} reports the training costs of the baseline and RSDNet on the nuScenes dataset. We suggest that the future works can leverage some knowledge distillation techniques to further reduce the training costs.
    \item \textbf{Limited Robustness to Noise.} Since RSDNet only adopts DDPMs as an auxiliary head, this can inherit the DDPM training paradigm only in an indirect manner. Moreover, the noise samples and targets generated in the latent feature space deviate from the real data distribution, thus lacking important details of the real samples. This means that the noise robustness of RSDNet is weaker than that of directly using DDPMs in the conventional manner (\ie, the score matching).
    \item \textbf{Difficult to Apply to Generative Tasks} The purpose of DLF is to overcome the limitations of multi-step inference while modeling multiple perturbations, lowering the barrier for applying DDPMs to 3D perception tasks. DLF treats DDPMs as an auxiliary denoising autoencoding branch, enhancing the reconstruction and generalization scene context for the model. However, this design makes the unsuitable for generative tasks, as the task result is determined by the main backbone and the denoising network is detached during inference. Therefore, we recommend adopting the traditional score matching paradigm when applying DDPMs to  generative tasks.

\end{itemize}

\section{More Visualization Results}

\subsection{More Multi-type Noise Samples and Targets.} 
\begin{figure*}[htp]
	\centering
\includegraphics[width=\textwidth]
 {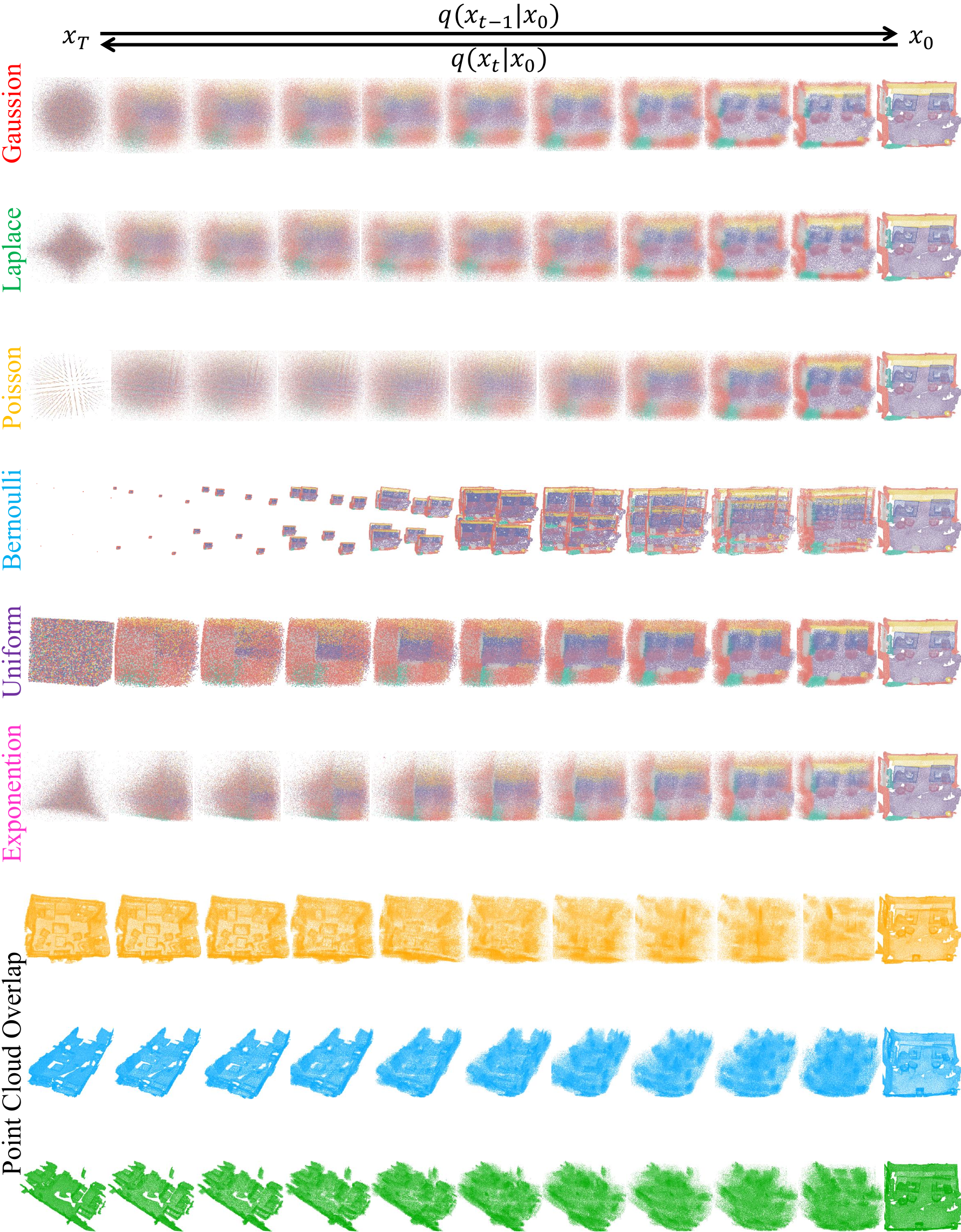}

\caption{Visualizations of multi-type noisy samples and targets. According to the sample fitting rule (noising via $q(\bm{x_t}|\bm{x_0})$, denoising via $q(\bm{x_{t-1}}|\bm{x_0})$), this can theoretically construct sample fitting targets for any distribution.}
	\label{supp_fig5}
\end{figure*}

Fig.~\ref{supp_fig5} illustrates more types of noise samples and targets constructed based on the sample fitting rules, which extend beyond those presented in the main text. Theoretically, this design aligns with the diffusion noising via $q(\bm{x_t}|\bm{x_0})$ and the sampling denoising through $q(\bm{x_{t-1}}|\bm{x_0})$. To generate these noise samples and targets, we simply alter the distribution of $\bm{\epsilon}$. For example, We can generate the Poisson-distributed noise sample and target through: $\bm{x_t}=\sqrt{\overline{\alpha}_t}\bm{x_0} + \sqrt{1-\overline{\alpha}_t}\bm{\epsilon_{t-1}}, \bm{\epsilon_{t-1}} \sim Possion(\lambda)$.

\subsection{More Detection Results.} 
\begin{figure*}[htp]
	\centering
\includegraphics[width=0.95\textwidth]
 {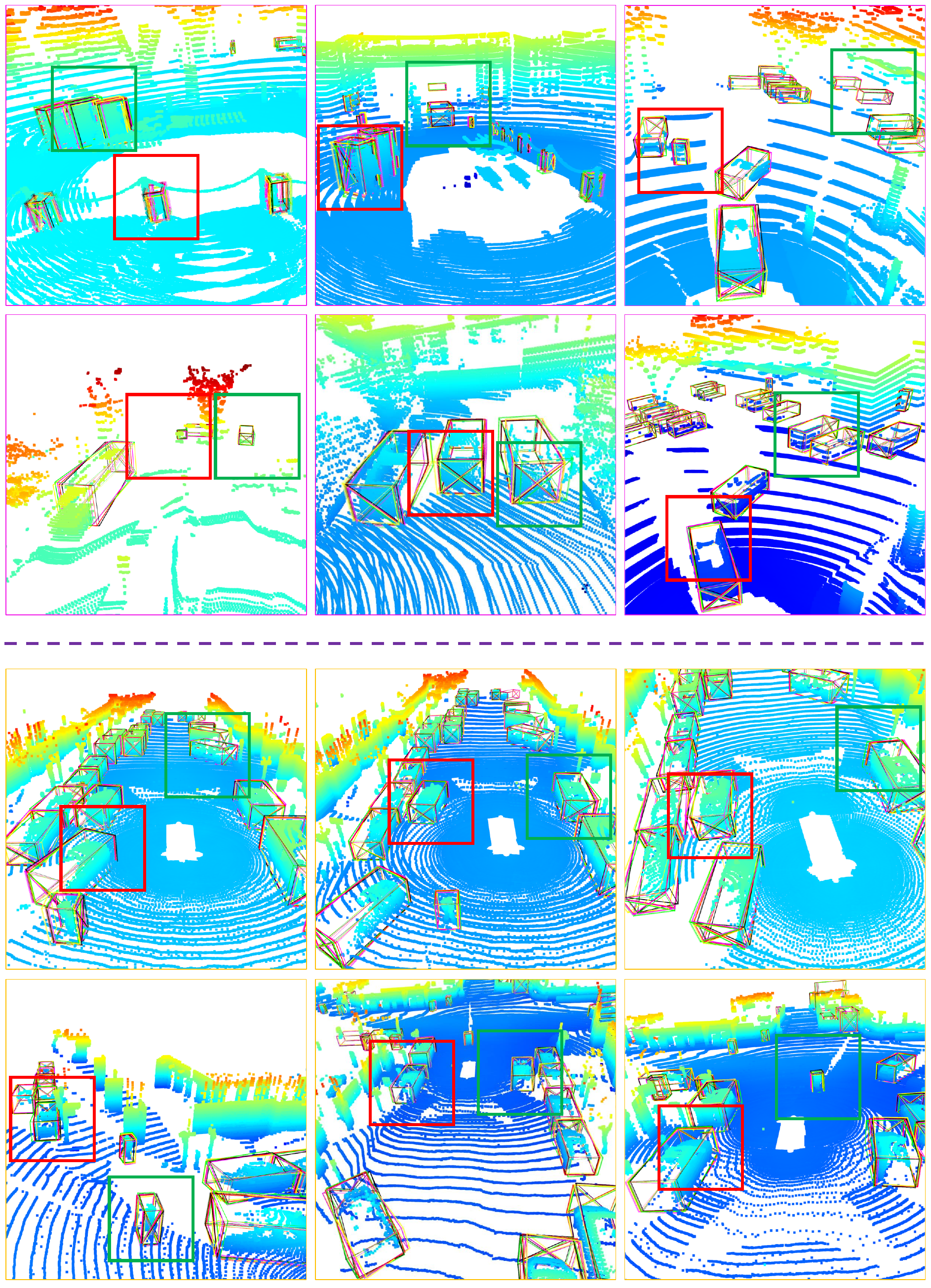}
\caption{More visualizations of detection results on nuScenes (Up) and Waymo Open (Down).}
	\label{supp_fig6}
\end{figure*}

We also present additional detection visualization results on nuScenes (up, the pink boxes) and Waymo (down, the orange boxes). In Fig.~\ref{supp_fig6}, the red, yellow, green, pink, and black boxes represent the Ground Truth, HEDNet, SAFDNet, FSHNet, and RSDNet, respectively. 

\newpage

\begin{table}[h]
        \scriptsize
	%\begin{center}
  \resizebox{0.475\textwidth}{!}{
	\begin{tabular}{p{1.3cm}p{0.8cm}p{0.8cm}p{0.005cm}p{1.0cm}p{1.5cm}p{1.5cm}}	
        \Xhline{1pt}

        Mehtod
        &\makecell[c]{NDS}
        &\makecell[c]{mAP}
        &\quad
        &\makecell[c]{$\#$Param}
        &\makecell[c]{MTT/MTM}
        &\makecell[c]{MIT/MIM}\\
       \hline
       
        Baseline
        &\makecell[c]{71.2}
        &\makecell[c]{68.0}
        &\quad
        &\makecell[c]{11.1M}
        &\makecell[c]{40h/10.5G}
        &\makecell[c]{0.16s/5.8G}\\

        \rowcolor{gray!20}
        RSDNet
        &\makecell[c]{71.9}
        &\makecell[c]{68.9}
        &\quad
        &\makecell[c]{16.5M}
        &\makecell[c]{43h/11.2G}
        &\makecell[c]{0.16s/5.8G}\\

        \Xhline{1pt}
        
	\end{tabular}
	%\end{center}
 }
 \vspace{-5pt}
	\caption{The comparison between RSDNet and Baseline for the performance and the computational cost. 'MTT' means the \textit{Mean Training Time}. 'MTM' presents the \textit{Mean Training  Memory}. `MIT' and `MIM' indicate the \textit{Mean Inference Time} and the \textit{Mean Inference Memory} for each point cloud.}
	\label{supp_tab511}

\end{table}

\begin{table}[h]
        \scriptsize
	%\begin{center}
  \resizebox{0.48\textwidth}{!}{
	\begin{tabular}{p{3.0cm}p{3.5cm}}	
       \Xhline{1pt}
  
        Config (DUNet)
        &\makecell[c]{Parameter} \\
        \cline{1-2}

        **Architecture**
        &\makecell[c]{ } \\
        
        DM Input
        &\makecell[c]{$f_t^*(\bm{x_t})$} \\
        
        DM Target
        &\makecell[c]{$h_t^*(\bm{\epsilon})$} \\

        DM Input Channel 3D
        &\makecell[c]{64} \\

        DM Input Channel 2D
        &\makecell[c]{128} \\

        DM Output Channel
        &\makecell[c]{8} \\

        T
        &\makecell[c]{1000} \\

        T Dim
        &\makecell[c]{128} \\

        Beta Start
        &\makecell[c]{0.0001} \\

        Beta End
        &\makecell[c]{0.02} \\

        Noise Schedule
        &\makecell[c]{linear} \\

        Skip Connection Mode
        &\makecell[c]{add} \\

        Skip Connection Scale
        &\makecell[c]{equal} \\

        Noise Schedule
        &\makecell[c]{linear} \\
        \cline{1-2}
        
        **NCM**
        &\makecell[c]{ } \\
        
        Translation 2D
        &\makecell[c]{[-5.0, 0.0]} \\

        Scale 2D
        &\makecell[c]{[0.01, 1.0]} \\

        Rotation 2D
        &\makecell[c]{[-3.1415,0.0]} \\

        Translation 3D
        &\makecell[c]{[0.0, 5.0]} \\

        Scale 2D
        &\makecell[c]{[1.0, 2.0]} \\

        Rotation 2D
        &\makecell[c]{[0.0, 3.1415]} \\

        Data Target
        &\makecell[c]{$\bm{\epsilon}/\bm{x_0}$} \\
        \cline{1-2}

        **SGCL**
        &\makecell[c]{ } \\

        Class Num
        &\makecell[c]{10} \\   

        Point Cloud Range
        &\makecell[c]{[-54.0, -54.0, -5.0, 54.0, 54.0, 3.0]} \\           

        Voxel Size
        &\makecell[c]{[0.3, 0.3, 0.2]} \\  

        Semantic Hidden Dim
        &\makecell[c]{128} \\  

        Geometric Hidden Dim
        &\makecell[c]{128} \\  

        Con Skip Connection Mode
        &\makecell[c]{add} \\  
        
        \Xhline{1pt}
        
	\end{tabular}
	%\end{center}
}
	\caption{The parameters of network framework for DUNet.}
	\label{supp_tab311}

\end{table}

\newpage

\begin{table}[h]
        \scriptsize
	%\begin{center}
  \resizebox{0.48\textwidth}{!}{
	\begin{tabular}{p{3.0cm}p{3.5cm}}	
       \Xhline{1pt}
  
        Config (nuScenes)
        &\makecell[c]{Parameter} \\
        \cline{1-2}

        **Architecture**
        &\makecell[c]{TransFusion} \\
        \cline{1-2}

        **3D Sparse Backbone**
        &\makecell[c]{ } \\

        VFE
        &\makecell[c]{DynamicVoxelVFE} \\

        With Distance
        &\makecell[c]{False} \\

        Use Norm
        &\makecell[c]{True} \\

        Use Absolute XYZ
        &\makecell[c]{True} \\

        Num Filters
        &\makecell[c]{[64,64]} \\

        Feature DIM
        &\makecell[c]{128} \\

        Win Size
        &\makecell[c]{20} \\
        \cline{1-2}
        
        **2D Sparse Backbone**
        &\makecell[c]{ } \\

        AFD Feature DIM
        &\makecell[c]{128} \\

        AFD Num Layers
        &\makecell[c]{4} \\

        AFD Num SBB
        &\makecell[c]{[2, 1, 1]} \\

        AFD Down Stride
        &\makecell[c]{[1, 2, 2]} \\

        AFD Down Kernal Size
        &\makecell[c]{[3, 3, 3]} \\

        FG Threshold
        &\makecell[c]{0.2} \\        

        Featmap Stride
        &\makecell[c]{2} \\   

        Group Pooling Kernel Size
        &\makecell[c]{[9, 15, 5, 5]}\\
        \cline{1-2}
        
        **Sparse Detection Head**
        &\makecell[c]{TransFusionHead} \\

        Query Radius
        &\makecell[c]{2} \\

        Query Local
        &\makecell[c]{True} \\

        Input Features
        &\makecell[c]{128} \\

        Num Proposals
        &\makecell[c]{600} \\

        Hidden Channel
        &\makecell[c]{128} \\

        Num Heads
        &\makecell[c]{8} \\

        NMS Kernel Size
        &\makecell[c]{3} \\

        FFN Channel
        &\makecell[c]{256} \\

        Dropout
        &\makecell[c]{0.1} \\

        BN Momentum
        &\makecell[c]{0.1} \\
        \cline{1-2}

        **Optimization**
        &\makecell[c]{ } \\        

        Batch Size
        &\makecell[c]{8} \\  

        Num Epoch
        &\makecell[c]{36} \\ 

        Optimizer
        &\makecell[c]{Adam} \\ 

        Scheduler
        &\makecell[c]{Cosine} \\ 

        Learning Rate
        &\makecell[c]{0.004} \\ 

        Weight Decay
        &\makecell[c]{0.03} \\ 

        Momentum
        &\makecell[c]{0.9} \\ 

        Moms
        &\makecell[c]{[0.95, 0.85]} \\         

        Pct Start
        &\makecell[c]{0.6} \\    

        Decay Step List
        &\makecell[c]{[35, 45]} \\    

        LR Decay
        &\makecell[c]{0.1} \\    
        \cline{1-2}

        **Loss**
        &\makecell[c]{ } \\  

        3D MSE Weight
        &\makecell[c]{0.1} \\    

        2D MSE Weight
        &\makecell[c]{0.1} \\   

        Regression Weight
        &\makecell[c]{0.25} \\   

        HM Weight
        &\makecell[c]{1.0} \\   

        IOU Weight
        &\makecell[c]{0.5} \\   

        IOU Regression Weight
        &\makecell[c]{0.5} \\  

        Class Weight
        &\makecell[c]{1.0} \\   
        
        \Xhline{1pt}
        
	\end{tabular}
	%\end{center}
}
	\caption{The parameters of network framework for RSDNet on the nuScenes dataset.}
	\label{supp_tab312}

\end{table}

\begin{table}[h]
        \scriptsize
	%\begin{center}
  \resizebox{0.48\textwidth}{!}{
	\begin{tabular}{p{3.0cm}p{3.5cm}}	
       \Xhline{1pt}
  
        Config (Waymo Open)
        &\makecell[c]{Parameter} \\
        \cline{1-2}

        **Architecture**
        &\makecell[c]{CenterPoint} \\
        \cline{1-2}

        **3D Sparse Backbone**
        &\makecell[c]{ } \\

        VFE
        &\makecell[c]{DynamicVoxelVFE} \\

        With Distance
        &\makecell[c]{False} \\

        Use Norm
        &\makecell[c]{True} \\

        Use Absolute XYZ
        &\makecell[c]{True} \\

        Num Filters
        &\makecell[c]{[64,64]} \\

        Feature DIM
        &\makecell[c]{128} \\

        Win Size
        &\makecell[c]{12} \\
        \cline{1-2}
        
        **2D Sparse Backbone**
        &\makecell[c]{ } \\

        AFD Feature DIM
        &\makecell[c]{128} \\

        AFD Num Layers
        &\makecell[c]{1} \\

        AFD Num SBB
        &\makecell[c]{[8, 4, 4]} \\

        AFD Down Stride
        &\makecell[c]{[1, 2, 2]} \\

        AFD Down Kernal Size
        &\makecell[c]{[3, 3, 3]} \\

        FG Threshold
        &\makecell[c]{0.4} \\        

        Featmap Stride
        &\makecell[c]{2} \\   

        Group Pooling Kernel Size
        &\makecell[c]{[7, 3, 3]}\\
        \cline{1-2}
        
        **Sparse Detection Head**
        &\makecell[c]{SparseDynamicHead} \\

        Input Features
        &\makecell[c]{128} \\

        Head Conv Type
        &\makecell[c]{spconv} \\

        Num HM Conv
        &\makecell[c]{2} \\

        BN EPS
        &\makecell[c]{0.001} \\

        BN MOM
        &\makecell[c]{0.01} \\

        R Factor
        &\makecell[c]{0.5} \\

        Dynamic Pos Num
        &\makecell[c]{5} \\

        DCLA REG Weight
        &\makecell[c]{3} \\

        \quad
        &\makecell[c]{\quad} \\

        \quad
        &\makecell[c]{\quad} \\        
        
        \cline{1-2}

        **Optimization**
        &\makecell[c]{ } \\        

        Batch Size
        &\makecell[c]{16} \\  

        Num Epoch
        &\makecell[c]{12} \\ 

        Optimizer
        &\makecell[c]{Adam} \\ 

        Scheduler
        &\makecell[c]{Cosine} \\ 

        Learning Rate
        &\makecell[c]{0.002} \\ 

        Weight Decay
        &\makecell[c]{0.05} \\ 

        Momentum
        &\makecell[c]{0.9} \\ 

        Moms
        &\makecell[c]{[0.95, 0.85]} \\         

        Pct Start
        &\makecell[c]{0.5} \\    

        Decay Step List
        &\makecell[c]{[35, 45]} \\    

        LR Decay
        &\makecell[c]{0.1} \\    
        \cline{1-2}

        **Loss**
        &\makecell[c]{ } \\  

        3D MSE Weight
        &\makecell[c]{0.1} \\    

        2D MSE Weight
        &\makecell[c]{0.1} \\   

        Regression Weight
        &\makecell[c]{3.0} \\   

        Class Weight
        &\makecell[c]{1.0} \\   

        \quad
        &\makecell[c]{\quad} \\

        \quad
        &\makecell[c]{\quad} \\    

        \quad
        &\makecell[c]{\quad} \\    
        
        \Xhline{1pt}
        
	\end{tabular}
	%\end{center}
}
	\caption{The parameters of network framework for RSDNet on the Waymo Open dataset.}
	\label{supp_tab313}

\end{table}

\newpage

\end{document}